\documentclass{article}

\PassOptionsToPackage{numbers, compress}{natbib}
\usepackage{booktabs,tabularx, colortbl} 
\usepackage{multirow}
\usepackage{float}
\usepackage{amsfonts} 
\usepackage{bm}
\usepackage{mathtools}
\usepackage{amsmath}
\usepackage{algorithm}
\usepackage{algorithmic}
\usepackage{subcaption}

\usepackage{xstring}

\usepackage{amssymb}
\usepackage{amsthm}

\usepackage{wrapfig}
\usepackage{booktabs}
\usepackage{multirow}
\usepackage{graphicx}
\usepackage{colortbl}
\usepackage{array}
\usepackage{enumitem}

\definecolor{lightred}{RGB}{255, 230, 230}
\definecolor{lightgreen}{RGB}{230, 255, 230}

\usepackage[most]{tcolorbox}

\usepackage{tabularx}
\usepackage{colortbl} 
\definecolor{s-gray}{gray}{0.95} 

\definecolor{ggreen}{rgb}{0.0, 0.6, 0.0}
\definecolor{rred}{rgb}{0.75, 0.0, 0.0}
\definecolor{bblue}{rgb}{0.13, 0.67, 0.8}

\definecolor{BoxBackground}{RGB}{240, 240, 240} 
\definecolor{BoxFrame}{RGB}{0, 0, 0} 
\definecolor{TitleBackground}{RGB}{0, 0, 0} 
\definecolor{TitleText}{RGB}{255, 255, 255} 
\definecolor{deepgreen}{RGB}{0,100,0}
\tcbset{
  academicbox/.style={
    boxsep=5pt,
    left=2pt,
    right=2pt,
    bottom=0.5pt,
    boxrule=0.5pt,
    colback=BoxBackground,
    colframe=BoxFrame,
    colbacktitle=TitleBackground,
    coltitle=TitleText,
    enhanced,
    attach boxed title to top left={yshift=-0.1in,xshift=0.1in},
    boxed title style={boxrule=0pt,colframe=white},
    title={#1},
  }
}

\newtcolorbox{AcademicBox}[1][]{academicbox=#1}
\definecolor{SoftBlue}{RGB}{135, 206, 250}  
\definecolor{SoftOrange}{RGB}{255, 224, 178} 
\definecolor{SoftGreen}{RGB}{144, 238, 144}  
\definecolor{CorrectGreen}{RGB}{76, 175, 80} 
\definecolor{ErrorRed}{RGB}{211, 47, 47} 

\usepackage[main, final]{neurips_2026}

\usepackage[utf8]{inputenc} 
\usepackage[T1]{fontenc}    
\usepackage{hyperref}       
\usepackage{url}            
\usepackage{booktabs}       
\usepackage{amsfonts}       
\usepackage{nicefrac}       
\usepackage{microtype}      
\usepackage{xcolor}         
\usepackage{cleveref}
\title{Flux Attention: Context-Aware Hybrid Attention for Efficient LLMs Inference}

\author{%
  Quantong Qiu\textsuperscript{1} \quad Zhiyi Hong\textsuperscript{1} \quad Yi Yang\textsuperscript{1} \quad Haitian Wang\textsuperscript{1} \\[0.2em]
  \textbf{Kebin Liu\textsuperscript{2}} \quad \textbf{Qingqing Dang\textsuperscript{2}} \quad \textbf{Juntao Li\textsuperscript{1,\dag}} \quad \textbf{Min Zhang\textsuperscript{1}} \\[0.6em]
  \textsuperscript{1}Soochow University, China \qquad \textsuperscript{2}Baidu Inc., China \\
  \textsuperscript{\dag}Corresponding author: \texttt{ljt@suda.edu.cn} \\[0.4em]
  Code: \url{https://github.com/qqtang-code/FluxAttention}
}

\begin{document}

\maketitle

\begin{abstract}
The quadratic computational complexity of standard attention mechanisms presents a severe scalability bottleneck for LLMs in long-context scenarios.
While hybrid attention mechanisms combining Full Attention (FA) and Sparse Attention (SA) offer a potential solution, existing methods typically rely on static allocation ratios that fail to accommodate the variable retrieval demands of different tasks.
Furthermore, head-level dynamic sparsity often introduces severe computational load imbalance and synchronization long-tails, which hinder hardware acceleration during autoregressive decoding.
To bridge this gap, we introduce \emph{\textbf{Flux Attention}}, a context-aware framework that dynamically optimizes attention computation at the layer level.
By integrating a lightweight \emph{\textbf{Layer Router}} into frozen pretrained LLMs, the proposed method adaptively routes each layer to FA or SA based on the input context.
This layer-wise routing preserves high-fidelity information retrieval while ensuring contiguous memory access, translating theoretical computational reductions into practical wall-clock speedups.
As a parameter-efficient approach, our framework requires only 12 hours of training on 8$\times$A800 GPUs.
Extensive experiments across multiple long-context and mathematical reasoning benchmarks demonstrate that Flux Attention achieves a superior trade-off between performance and inference speed compared with baseline models, with speed improvements of up to $2.8\times$ and $2.0\times$ in the prefill and decode stages.
\end{abstract}
\section{Introduction}
\label{sec:intro}
\begin{figure}[ht!]
    \centering
    \begin{subfigure}{0.48\linewidth}
        \centering
        \includegraphics[width=\linewidth]{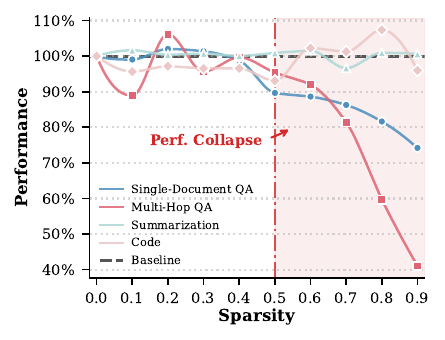}
        \caption{Performance vs. sparsity.}
    \end{subfigure}
    \hfill
    \begin{subfigure}{0.48\linewidth}
        \centering
        \includegraphics[width=\linewidth]{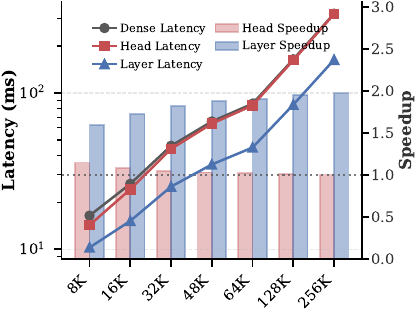}
        \caption{Decoding latency and speedup.}
    \end{subfigure}
    \caption{Impact of sparsity on performance and decoding efficiency. (a) Certain tasks suffer performance collapse beyond a specific threshold; this sweep uses Streaming Sparse Attention (SSA) with the configuration of Appendix~\ref{app:sparsity_config}. (b) Layer-level sparsity achieves substantial decoding speedup, while head-level sparsity yields marginal speedup.}
    \label{fig:preliminary}
\end{figure}

Large Language Models (LLMs) have demonstrated strong capabilities in handling extended context windows for tasks such as document analysis, long-form reasoning, and question answering~\citep{liu2025comprehensive, mei2025survey}. However, the standard Full Attention (FA) mechanism~\citep{vaswani2017attention} scales quadratically with sequence length, creating severe memory and computational bottlenecks during prefilling and autoregressive decoding. Sparse Attention (SA) mechanism addresses this by restricting computations to a subset of tokens to reduce the memory footprint~\citep{child2019generating, zaheer2020big}.

Modern architectures frequently employ hybrid attention mechanisms that integrate both FA and SA within a single network to balance inference efficiency and generation quality~\citep{zhang2025efficient}. Conventional hybrid models typically rely on a static allocation of dense and sparse computation. However, downstream applications exhibit highly varied computational demands, as detailed in our preliminary study (\Cref{subsec:observations}). Retrieval-intensive tasks require dense token interactions to locate specific information, whereas context-holistic tasks focus on overarching semantics and remain stable under high sparsity~\citep{peng2025accelerating}. Consequently, a static configuration risks performance degradation on retrieval tasks and wastes valuable computational resources on holistic tasks.

To achieve dynamic allocation, recent works~\citep{tang2026elasticattentiontesttimeadaptive} have explored fine-grained routing at the head level by assigning varying sparsity ratios to individual attention heads based on the input. While algorithmically flexible, this fine-grained routing introduces severe hardware inefficiencies during the memory-bandwidth-bound decode phase. Varying context lengths across heads lead to heterogeneous computational workloads within the same layer. This forces thread blocks executing sparse heads to idle while waiting for retrieval heads, creating a synchronization long-tail that prevents theoretical FLOP reductions from translating into actual wall-clock decoding speedups.

To overcome these challenges, we propose Flux Attention, a context-aware framework that dynamically optimizes attention computation at the layer level. Instead of managing individual heads, we introduce a lightweight Layer Router. By evaluating the semantic context of the input prompt, the router infers the underlying task demands and adaptively assigns each layer to FA or SA mode. This coarse granularity inherently preserves contiguous memory access, enabling the GPU to completely bypass the memory-intensive loading of historical KV tensors when SA is selected.

During training, we freeze all backbone LLM parameters and update only the lightweight Layer Router module. We employ a Gumbel-Softmax~\citep{jang2016categorical} relaxation for differentiable soft routing, allowing the model to smoothly learn the correlation between context complexity and computational budget. During inference, this soft formulation is discretized into deterministic hard routing, successfully translating theoretical computational savings into substantial wall-clock speedups.

Extensive evaluations on models such as Qwen-3~\citep{yang2025qwen3technicalreport} and Llama-3.1~\citep{grattafiori2024llama} demonstrate that Flux Attention successfully adapts sparsity levels across diverse tasks. Our parameter-efficient training converges in just 12 hours on an 8-GPU A800 node. Flux Attention achieves a superior performance-efficiency trade-off compared to existing baselines, delivering up to a 2.7x speedup during the prefill phase and a 2.0x acceleration during autoregressive decoding.
\section{Preliminary}
\label{sec:Preliminary}

\subsection{Functional Heterogeneity in Attention Mechanisms}
\label{subsec:functional_heterogeneity}

During long-context inference, attention mechanisms in Large Language Models (LLMs) specialize functionally based on their sensitivity to historical context and computational demands. Specialized retrieval heads are essential for high-fidelity information recovery, as they precisely locate relevant tokens across extensive sequences~\citep{wu2024retrieval}. UnComp~\citep{UnComp} observe that heads with abnormally high entropy tend to aggregate at specific model depths to capture long-range dependencies. Layers dominated by these heads function as retrieval layers. To ensure precise retrieval, they require a Full Attention (FA) mode, where the Query ($Q$) interacts with all historical states Key ($K$) and Value ($V$):
\begin{equation}
\mathcal{O}_{r} = \text{Softmax} \left( Q K^\top \right) V,
\label{equ:full_attention}
\end{equation}
where the scaling factor is omitted for clarity. While FA preserves the complete context, its computational complexity is quadratic with sequence length $N$, posing challenges for efficient inference.

A substantial portion of heads instead focus on local semantic structures and are robust to context truncation. Layers predominantly composed of these sparse heads operate as sparse layers. Sparse layers employ a Sparse Attention (SA) mechanism to reduce computational overhead in long-sequence processing. SA optimizes efficiency by performing attention operations on a condensed subset of the most critical historical elements ($\tilde{K}$ and $\tilde{V}$):
\begin{equation}
\mathcal{O}_{s} = \text{Softmax} \left( Q \tilde{K}^\top \right) \tilde{V}.
\label{equ:sparse_attention}
\end{equation}

\subsection{Rethinking Hybrid Attention Mechanisms}
\label{subsec:rethinking}

To balance generation quality and inference efficiency, various hybrid attention mechanisms have been proposed. Existing methods, such as PruLong~\citep{bhaskar2025cache}, DuoAttention~\citep{duoattention}, and LycheeDecode~\citep{lin2026lycheedecode}, adopt a static allocation strategy. They identify retrieval heads offline and permanently assign them full historical states, while uniformly sparsifying the context for the remaining heads across all tasks. 

However, the demand for precise information retrieval varies depending on the specific task and input prompt. Elastic Attention~\citep{tang2026elasticattentiontesttimeadaptive} suggests dynamic, context-aware sparsity at the head level, which adjusts the retention of historical states dynamically. Although this fine-grained allocation optimizes the theoretical efficiency-performance trade-off, it yields limited actual decoding acceleration. The dynamic adjustment at the head level introduces significant system-level overhead and irregular memory access patterns during deployment, limiting the achievable speedup during the decode phase.

\subsection{Motivational Observations}
\label{subsec:observations}

To investigate the limitations of existing sparsity mechanisms, we formalize the quantification of model-level sparsity. The Model Sparsity Ratio ($\Omega_{\mathrm{MSR}}$) quantifies the overall proportion of sparse attention mechanisms applied across the model:
\begin{equation}
\Omega_{\mathrm{MSR}}
= \frac{1}{H \times L}
\sum_{\ell=1}^{L}\sum_{h=1}^{H}
\mathbb{I}\!\left[\pi^{(\ell, h)}=\mathrm{SA}\right],
\label{equ:msr}
\end{equation}
where $\pi^{(\ell, h)}$ denotes the assigned attention mode (FA or SA) for head $h$ in layer $\ell$, and $\mathbb{I}[\cdot]$ is the indicator function.

\paragraph{Settings} 
To investigate the impact of varying sparsity ratios ($\Omega_{\mathrm{MSR}}$) on long-context LLMs, we profile task accuracy and decode latency.
For the accuracy evaluation in Figure~\ref{fig:preliminary}(a), we use a matrix entropy metric based on UnComp~\citep{UnComp} to quantify the information density of individual layers.
We rank the layers using these calculated entropy scores and progressively replace the lowest-scoring ones with SA.
Model performance is then evaluated across real-world tasks from LongBench~\citep{bai2024longbench}.
For hardware efficiency (Figure~\ref{fig:preliminary}(b)), we compare the decode latency and achievable speedup of our layer-level sparsity against a static head-level sparsity baseline.
Appendix~\ref{appendix:layer_sparsification_settings} provides details on the entropy scoring formulation and latency measurement implementation.

\paragraph{Results} 
Our analysis reveals two bottlenecks in current hybrid attention mechanisms.
First, as shown in Figure~\ref{fig:preliminary}(a), model performance does not degrade linearly with increasing $\Omega_{\mathrm{MSR}}$.
Instead, accuracy drops sharply for retrieval-intensive tasks once a specific sparsity threshold is exceeded.
This indicates that static sparsity assignments do not adapt to varying contextual demands, necessitating a context-aware dynamic retention strategy for historical states.
Second, Figure~\ref{fig:preliminary}(b) demonstrates a distinct discrepancy in hardware efficiency.
While head-level sparsity provides algorithmic flexibility, it introduces severe hardware bottlenecks during the memory-bandwidth-bound decode phase.
It creates a severe synchronization long-tail effect. Thread blocks executing sparse heads finish quickly but must idle while waiting for memory-intensive retrieval heads within the same layer. 
This intra-layer load imbalance yields only marginal wall-clock speedups. 
In contrast, layer-level sparsity ensures uniform computational workloads across all thread blocks. By completely bypassing historical KV loading for designated layers, it eliminates synchronization stalls, effectively translating theoretical FLOP reductions into substantial decode acceleration.

These observations present a fundamental dilemma. \textit{Fine-grained head-level sparsity is hardware-unfriendly during decode, whereas static sparsity risks performance collapse}. To address this, we propose a dynamic, context-aware hybrid attention mechanism operating at the layer level to balance model performance with inference efficiency.

\section{Methodology}
\label{sec:method}
\begin{figure}[t]
    \centering
    \includegraphics[width=\linewidth]{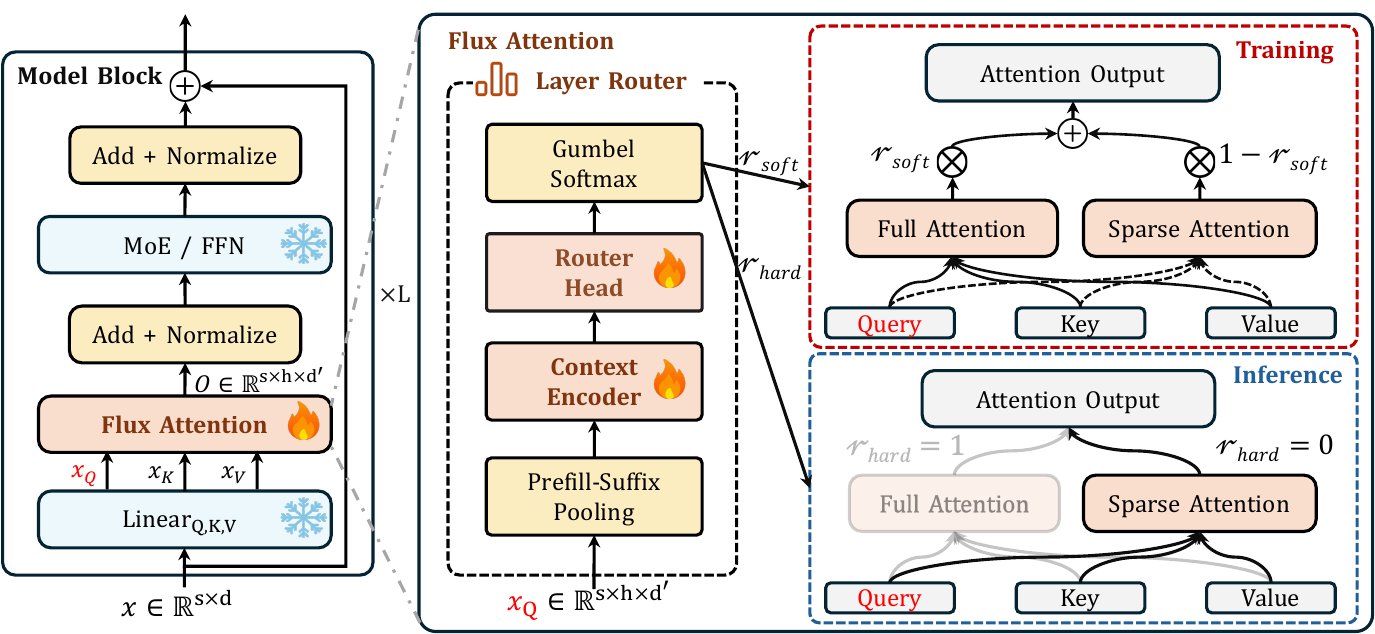}
    \caption{Overview of our dynamic layer-level routing architecture. The model incorporates a Layer Router that assigns each layer to either FA or SA based on the input query $x_Q$.}
    \label{fig:arc}
\end{figure}

We introduce a Flux Attention mechanism to address the hardware inefficiencies of fine-grained sparsity and the rigidity of static allocations. As illustrated in Figure~\ref{fig:arc}, our architecture relies on a dynamic Layer Router that adaptively assigns each layer to either FA or SA based on the input query. This approach is parameter-efficient: the original LLM backbone parameters remain strictly frozen during training. Optimization only updates the lightweight components of the Layer Router, ensuring rapid convergence while preserving pre-trained weights.

\subsection{Context-Aware Layer Router Design}
\label{subsec:layer_router}

Within the Flux Attention module, a lightweight Layer Router determines the optimal attention mechanism for a given context. 

\paragraph{Architecture and Feature Extraction}
As shown in Figure~\ref{fig:arc}, the router receives the incoming query tensor $x_Q \in \mathbb{R}^{s \times h \times d'}$ as input, where $s$ represents the sequence length, $h$ denotes the number of heads, and $d'$ indicates the head dimension.
To efficiently extract semantic context, we apply a Prefill-Suffix Pooling operation to $x_Q$ to extract representations of the initial and final prompt tokens. This operation efficiently aggregates the token-level features into a single sequence-level descriptor. 
Subsequently, a Context Encoder (MLP) processes this pooled representation to capture contextual dependencies, after which a Router Head (MLP) projects these features into unnormalized routing logits, denoted as $\pi_{\text{FA}}$ and $\pi_{\text{SA}}$.

\paragraph{Differentiable Training via Soft Routing}
Optimizing the Layer Router is challenging because the binary routing decisions are discrete and non-differentiable.To address this, we apply the Gumbel-Softmax relaxation~\citep{jang2016categorical} to enable end-to-end backpropagation.
During training, we sample continuous routing weights $r_{\text{soft}} \in (0, 1)$ , which represent the probability of selecting the FA mechanism. This computation is defined as follows:
\begin{equation}
r_{\text{soft}} = \frac{\exp((\pi_{\text{FA}} + g_{\text{FA}})/\tau)}{\exp((\pi_{\text{FA}} + g_{\text{FA}})/\tau) + \exp((\pi_{\text{SA}} + g_{\text{SA}})/\tau)}
\label{equ:gumbel}
\end{equation}
where $g_{\text{FA}}, g_{\text{SA}} \sim \text{Gumbel}(0, 1)$ are independent and identically distributed samples drawn from the Gumbel distribution, and $\tau > 0$ denotes the temperature parameter. The output of the Flux Attention layer is then computed as a convex combination:
\begin{equation}
\mathcal{O}_{\text{train}} = r_{\text{soft}} \cdot \text{FA}(Q, K, V) + (1 - r_{\text{soft}}) \cdot \text{SA}(Q, \tilde{K}, \tilde{V}).
\label{equ:soft_routing}
\end{equation}

The temperature $\tau$ controls the smoothness of the routing distribution. We employ a temperature annealing schedule to minimize the train-test discrepancy. Initially, $\tau$ is set to a high value to encourage exploration and ensure smooth gradient flow. As training progresses, $\tau$ linearly decays towards a small value.

\paragraph{Deterministic Inference via Hard Routing}
During the inference phase, the router outputs a binary decision $r_{\text{hard}} \in \{0, 1\}$ using an $\arg\max$ operation over the generated logits.
When $r_{\text{hard}} = 0$, the layer executes the SA mechanism.

\subsection{Training Objective and Sparsity Constraint}
\label{subsec:loss_function}
We formulate the training objective as a constrained optimization problem to balance generation quality and computational efficiency. Without intervention, the router tends to degenerate by sending all queries to the FA mode, which trivially minimizes the language modeling loss. 

A dynamic penalty mechanism controls the inference budget. Let $\boldsymbol{t} \in (0, 1)$ denote the target computational budget for sparse computation (i.e., the permissible fraction of SA layers, corresponding to $\Omega_{\mathrm{MSR}}$).
Notably, instead of enforcing a rigidly fixed $\boldsymbol{t}$ for each task, we impose \emph{task-dependent non-tight constraints} with predefined lower and upper bounds, since the optimal sparsity for a given task is inherently unknown. We therefore define the sparsity deviation as $L_{\mathrm{diff}}(\mathcal{X}) = \mathbb{E}_{\mathcal{X}}[1 - r_{\text{soft}}] - \boldsymbol{t}$, which represents the gap between the expected sparse routing probability across all layers and the allocated budget. We solve the overall optimization objective via Lagrangian relaxation:
\begin{equation}
\max_{\lambda_1, \lambda_2 \ge 0} \min_{\theta} \underbrace{L_{\mathrm{language}}(\mathcal{X})}_{\mathrm{language~modeling}} + \underbrace{\lambda_1 L_{\mathrm{diff}}(\mathcal{X}) + \lambda_2 L_{\mathrm{diff}}^{2}(\mathcal{X})}_{\mathrm{sparsity~regularization}},
\label{equ:total_loss}
\end{equation}
where $\theta$ represents the trainable parameters of the Layer Router, and $L_{\mathrm{language}}(\mathcal{X})$ is the standard cross-entropy loss. The Lagrangian multipliers $\lambda_1$ and $\lambda_2$ are task-specific trainable Lagrange multipliers optimized via gradient ascent~\citep{bhaskar2025cache}, which decouple the sparsity--performance trade-offs across tasks and mitigate optimization conflicts.

\subsection{Efficient Deployment}
\label{subsec:deployment_of_attn_router}

To translate theoretical sparsity gains into real-world inference acceleration and memory savings, Flux Attention decouples routing computation between the prefill and decode phases, with a sparse-decode implementation aligned with our experimental settings.
The Layer Router infers only once during the prefill phase, generating a deterministic hard routing decision ($r_{\text{hard}} \in \{0, 1\}$) per layer based on the input context. This decision is cached and reused across all decoding steps, eliminating per-token routing overhead. The realized model-level sparsity of a request is therefore measured directly from the cached decisions as $\Omega_{\mathrm{MSR}} = \frac{1}{L}\sum_{\ell=1}^{L}\left(1 - r^{(\ell)}_{\text{hard}}\right)$; we report the realized value throughout rather than the target $\boldsymbol{t}$.
Routing is thus context-aware rather than token-wise: recomputing it at every generated token would add control overhead and break the homogeneous per-layer execution that produces the systems benefit, while the underlying task intent usually remains stable within a single response. In multi-turn or agentic use the route is refreshed at the natural turn boundary, where the extended dialogue is re-prefilled and the router runs again on the enlarged prompt (Appendix~\ref{appdix:multiturn}); the cached assignment is fixed only within one decode trajectory, not globally.
Our sparse-decode configuration further optimizes efficiency: for sparse layers, we only maintain the minimal KV cache required by the sparse kernel, fully bypassing full historical KV access and storage; for retrieval layers, complete KV cache is retained to preserve retrieval performance. This design delivers significant decoding speedups and KV cache reduction in long-context scenarios.The detailed execution flow of the inference is summarized in Algorithm~\ref{alg:flux_inference}.

\begin{table*}[ht!]
\centering
\caption{Performance on LongBench-E~\citep{bai2024longbench}. We report average performance (Perf.) and $\Omega_{\mathrm{MSR}}$ per task category. 
The 1st and the 2nd performance in each comparison group are highlighted with \textbf{bold font} and \underline{underlined}, respectively. Gray-shaded rows denote the sparse-decode configuration.
}

\resizebox{\textwidth}{!}{
\begin{tabular}{l | cc | cc | cc | ccc | cc |cc | cc}
\toprule
\multirow{2}{*}{\textbf{Method}} & \multicolumn{2}{c|}{\textbf{S-Doc QA}} & \multicolumn{2}{c|}{\textbf{M-Doc QA}} & \multicolumn{2}{c|}{\textbf{Summ}} & \multicolumn{3}{c|}{\textbf{In-Context}} & \multicolumn{2}{c|}{\textbf{Synthetic}} & \multicolumn{2}{c|}{\textbf{Code}} & \multicolumn{2}{c}{\textbf{Avg.}} \\
\cmidrule(lr){2-3} \cmidrule(lr){4-5} \cmidrule(lr){6-7} \cmidrule(lr){8-10} \cmidrule(lr){11-12} \cmidrule(lr){13-14} \cmidrule(lr){15-16}
 & \rotatebox[origin=c]{0}{Qasper} & \rotatebox[origin=c]{0}{MF-en} & \rotatebox[origin=c]{0}{HotQA} & \rotatebox[origin=c]{0}{2Wiki} & \rotatebox[origin=c]{0}{Gov.} & \rotatebox[origin=c]{0}{M.News} & \rotatebox[origin=c]{0}{TREC} & \rotatebox[origin=c]{0}{TQA} & \rotatebox[origin=c]{0}{SAMS} & \rotatebox[origin=c]{0}{PCount} & \rotatebox[origin=c]{0}{PRe} & \rotatebox[origin=c]{0}{RB-P} & \rotatebox[origin=c]{0}{Lcc} & \textbf{Perf.} & $\mathbf{\Omega_{\mathrm{\bf MSR}}}$\\
\arrayrulecolor{black}\midrule

\rowcolor{blue!10}
\multicolumn{16}{c}{\textbf{Qwen3-4B backbone model}} \\
\arrayrulecolor{black!20}\midrule
Qwen3-4B & 35.21 & 52.16 & 44.81 & 32.15 & 33.47 & 23.45 & 70.67 & 88.22 & 39.74 & 2.33 & 96.84 & 50.84 & 57.93 & 48.45 & - \\
+ DuoAttention & \textbf{35.83} & 49.84 & 47.09 & 32.24 & \underline{33.32} & \textbf{23.70} & \textbf{69.33} & 85.87 & 39.75 & \textbf{4.50} & \underline{94.57} & 50.56 & 57.43 & 48.22 & 0.50 \\
+ PruLong & 34.15 & 50.78 & 44.48 & \underline{32.89} & 32.96 & 23.53 & 67.67 & \textbf{88.69} & 39.55 & 3.17 & 90.17 & 49.00 & 54.07 & 47.16 & 0.50 \\
+ TriangleMix & 35.55 & \textbf{52.02} & 45.37 & 31.76 & \underline{33.32} & \textbf{23.70} & \underline{69.00} & 88.20 & 39.74 & 3.83 & 91.51 & 48.58 & 56.38 & 47.72 & 0.50 \\
\arrayrulecolor{black!40}\midrule
+ FluxAttn (FA-SSA) & 35.02 & 49.44 & \underline{49.64} & 32.27 & 33.26 & 23.48 & \textbf{69.33} & \underline{88.29} & 39.78 & 1.50 & 94.56 & \textbf{53.44} & \underline{59.69} & \underline{48.72} & 0.44 \\
+ FluxAttn (FA-XA) & \underline{35.74} & \underline{51.70} & 45.83 & 32.34 & \textbf{33.57} & \underline{23.66} & \underline{69.00} & 87.23 & 39.81 & 3.50 & 93.74 & 50.81 & 59.28 & 48.32 & 0.53 \\
+ FluxAttn (FA-TA) & 35.02 & 50.89 & 45.17 & \textbf{34.24} & 33.02 & 23.53 & \underline{69.00} & 88.08 & \textbf{40.38} & \underline{3.94} & \textbf{96.06} & \underline{51.68} & \textbf{60.00} & \textbf{48.76} & 0.47 \\
\arrayrulecolor{black!40}\midrule
\rowcolor{black!15}
+ FluxAttn (FA-SSA) & 35.10 & 51.68 & \textbf{49.65} & 32.86 & 33.04 & 23.42 & \textbf{69.33} & 88.00 & \underline{40.00} & 1.67 & 94.47 & 51.40 & 58.68 & 48.59 & 0.44 \\

\arrayrulecolor{black}
\midrule
\rowcolor{blue!25}
\multicolumn{16}{c}{\textbf{Qwen3-8B backbone model}} \\
\arrayrulecolor{black!20}\midrule
Qwen3-8B & 41.22 & 49.92 & 58.98 & 44.21 & 33.27 & 23.42 & 71.33 & 86.77 & 41.83 & 2.00 & 98.33 & 56.08 & 66.31 & 52.16 & - \\
+ DuoAttention & \textbf{41.78} & \textbf{51.55} & 55.96 & 41.70 & 33.24 & 23.34 & 69.33 & \textbf{89.35} & 41.62 & 0.50 & 98.93 & 57.54 & \underline{69.39} & 52.13 & 0.50 \\
+ PruLong & 37.95 & 51.20 & 51.94 & 36.48 & 33.11 & 23.36 & 69.00 & 87.90 & \underline{42.11} & 1.00 & 98.00 & 57.05 & 67.66 & 50.80 & 0.50 \\
+ TriangleMix & 40.82 & \underline{51.31} & 57.57 & \textbf{44.51} & \textbf{33.32} & 23.35 & 71.33 & 86.73 & 41.79 & \textbf{2.00} & 94.33 & 55.04 & 65.89 & 51.65 & 0.50 \\
\arrayrulecolor{black!40}\midrule
+ FluxAttn~(FA-SSA) & 40.30 & 50.49 & 56.02 & 40.90 & 33.01 & \textbf{23.55} & \underline{71.67} & 88.31 & 41.61 & 0.33 & \textbf{100.00} & \textbf{59.46} & 68.27 & \underline{52.18} & 0.46 \\
+ FluxAttn~(FA-XA) & 40.41 & 50.26 & \underline{57.78} & 40.57 & \underline{33.27} & \underline{23.51} & 69.67 & 87.19 & \textbf{42.12} & \underline{1.33} & 99.33 & 55.41 & 65.51 & 51.57 & 0.51 \\
+ FluxAttn~(FA-TA) & \underline{41.00} & 49.76 & \textbf{58.19} & \underline{44.36} & \textbf{33.32} & 23.35 & 70.00 & \underline{88.77} & 41.70 & \underline{1.33} & \underline{99.67} & 55.60 & 67.22 & \textbf{52.22} & 0.47 \\
\arrayrulecolor{black!40}\midrule
\rowcolor{black!15}
+ FluxAttn~(FA-SSA) & 39.92 & 50.04 & 55.72 & 40.81 & 33.03 & 23.50 & \textbf{72.00} & 88.48 & 40.96 & 0.33 & 99.22 & \underline{58.57} & \textbf{69.46} & 52.05 & 0.46 \\
\arrayrulecolor{black}
\midrule
\rowcolor{cyan!20}
\multicolumn{16}{c}{\textbf{Llama-3.1-8B-Instruct backbone model}} \\
\arrayrulecolor{black!20}\midrule
Llama-3.1-8B-Instruct &44.06 & 53.44 & 59.62 & 44.08 & 34.50 & 26.02 & 71.00 & 90.54 & 42.94 & 12.67 & 99.33 & 47.78 & 63.85 & 53.28 & - \\
+ DuoAttention & 34.63 & 50.74 & 49.70 & 36.41 & 34.25 & 25.78 & 70.00 & 91.45 & 42.13 & 9.80 & 97.33 & \underline{53.59} & \textbf{68.55} & 52.11 & 0.50 \\
+ PruLong & 41.51 & 52.36 & 50.46 & 37.57 & 34.25 & 25.86 & 66.33 & 89.93 & 41.72 & 9.07 & 97.00 & \textbf{56.84} & \underline{66.23} & 51.68 & 0.50 \\
+ TriangleMix & \underline{45.10} & \textbf{54.60} & 56.67 & 41.88 & 34.09 & 25.51 & \underline{71.33} & 90.93 & \underline{42.63} & \underline{10.62} & 94.67 & 43.64 & 59.35 & 51.67 & 0.50 \\
\arrayrulecolor{black!40}\midrule
+ FluxAttn~(FA-SSA) & \textbf{45.25} & \underline{54.42} & 54.54 & 41.34 & \underline{34.54} & \textbf{26.16} & 68.33 & \textbf{91.91} & 42.17 & 9.00 & 97.67 & 47.74 & 65.35 & 52.28 & 0.51 \\
+ FluxAttn~(FA-XA) & 42.14 & 53.13 & \textbf{58.53} & \textbf{43.50} & \textbf{34.66} & \underline{26.06} & 70.67 & \underline{91.46} & \textbf{43.13} & 8.00 & \textbf{99.67} & 50.91 & 64.78 & \textbf{53.07} & 0.72 \\
+ FluxAttn~(FA-TA) & 44.77 & 54.12 & 57.35 & \underline{43.43} & 34.31 & 25.80 & \textbf{72.33} & 91.32 & 42.62 & 9.33 & 98.33 & 45.48 & 60.70 & \underline{52.42} & 0.62 \\
\arrayrulecolor{black!40}\midrule
\rowcolor{black!15}
+ FluxAttn~(FA-SSA) & 43.76 & 53.41 & \underline{57.36} & 39.43 & 32.96 & 25.63 & 70.33 & 91.27 & 42.20 & \textbf{11.00} & \underline{98.67} & 45.60 & 66.17 & 52.30 & 0.51 \\
\arrayrulecolor{black}\bottomrule
\end{tabular}
}
\vspace{-0.5em}
\label{tab:longbench_main}
\end{table*}

\section{Experiments}
\label{sec:experiment}

\subsection{Settings}
\label{subsec:exp_settings}

\paragraph{Training and Data}
\label{subsec:train_settings}
We select Qwen3 (4B and 8B)~\citep{yang2025qwen3technicalreport} and Llama-3.1-8B-Instruct~\citep{grattafiori2024llama} as the backbone LLMs. We construct the training dataset by combining five sources: ChatQA2-Long-SFT-data~\citep{xu2024chatqa}, MuSiQue~\citep{trivedi2022musique}, CoLT-132K~\citep{li2025aixcoder}, GovReport~\citep{huang-etal-2021-efficient}, and XSum~\citep{xsum-emnlp}. This dataset covers both retrieval-intensive tasks  (Single-Doc QA and Multihop QA) and context-holistic tasks  (code completion, summarization, and in-context learning). The resulting dataset spans sequence lengths ranging from 1K to 64K tokens, and contains approximately 0.74B tokens in total. For the context-holistic and retrieval-intensive task categories, we empirically set $\boldsymbol{t}=1.0$ and $\boldsymbol{t}=0.45$, respectively, as motivated by \Cref{subsec:observations}. We conduct the training process using eight A800 GPUs, and each run completes within 12 hours. We provide additional training details in the Appendix~\ref{appendix:implementation_details} and list the hyperparameters in the table~\ref{tab:combined_hyperparameters}.

\paragraph{Evaluation}
\label{subsec:baseline}
We compare our method with representative sparsity approaches: DuoAttention~\citep{xiaoduoattention2025}, PruLong~\citep{bhaskar2025cache}, and TriangleMix~\citep{TriangleMix}.
The computation modes for sparse layer attention include Streaming Sparse Attention (SSA)~\citep{streamingllm}, XAttention (XA)~\citep{xattention}, and Triangle Attention (TA).
The configurations for layer computation follow the format of ``\{Retrieval Layer mode\}-\{Sparse Layer mode\}'' (e.g., FA-SSA denotes the use of FA for retrieval layers and SSA for sparse layers). All evaluations are conducted using the \texttt{LOOM-Eval} framework~\citep{tang2025loom}.
Due to space limitations, additional results are provided in the Appendix~\ref{appdix:eval_res}.

\subsection{Evaluation Results}
\label{subsec:eval_res_main}

\paragraph{Real-world Long-context Tasks}
Table~\ref{tab:longbench_main} presents the evaluation results on LongBench-E~\citep{bai2024longbench}, a real-world long-context benchmark that comprises 13 tasks across 6 categories with varying context lengths.
FluxAttn maintains the performance of the model on long-context tasks while achieving substantial context compression.
Across the Qwen3 series, variants of FluxAttn frequently match or slightly exceed the average performance of the full attention baselines.
We further evaluate the effect of applying sparse attention during the decode phase, as indicated in the shaded rows.
The method remains competitive under sparse decode. On Qwen3-4B, the sparse-decode configuration achieves an average score of 48.59, which remains above the full attention baseline.
For Qwen3-8B and Llama-3.1-8B-Instruct, the average scores (52.05 and 52.30, respectively) demonstrate only a slight degradation compared to the standard dense decoding approach.

\paragraph{Length Extrapolation Capability Testing}
To further assess the ability of the models to handle extreme context lengths, we evaluated our method on the RULER benchmark~\citep{hsieh2024ruler}, which tests length extrapolation capabilities from 8K to 256K tokens. The results are summarized in Table~\ref{tab:ruler_lb2_result}.
Overall, FluxAttn demonstrates robust length extrapolation, maintaining information retrieval and reasoning capabilities even at the 256K context boundary, where many existing sparse attention baselines experience severe performance degradation.
Consistent with our findings in real-world tasks, we also observe that extending sparsity to the decode phase (shaded rows) preserves the extrapolation capabilities.
The sparse-decode configuration of FluxAttn on Qwen3-4B achieves an average score of 67.19 (the highest among all methods in the comparison group) and a score of 56.00 at 256K. This result further validates that our method can achieve comprehensive efficiency gains without compromising ultra-long context understanding.

\paragraph{Long-form Reasoning and Math Tasks}
We further evaluate our models on the long-context reasoning benchmark LongBench-V2~\citep{bai2024longbench2}, as well as the mathematical reasoning tasks GSM8K~\citep{cobbe2021trainingverifierssolvemath} and AIME24~\citep{aime2024}. Table~\ref{tab:ruler_lb2_result} demonstrates that FluxAttn exhibits strong performance across both domains. On LongBench-V2, the proposed method attains the highest scores on both the easy and hard subsets among all baselines.
Furthermore, our approach improves the performance on the mathematical benchmarks, yielding the best results on GSM8K and AIME24.
This proves that FluxAttn robustly preserves complex logical reasoning capabilities.

\begin{table*}[t]
\centering
\caption{Model performance on RULER~\citep{hsieh2024ruler}, LongBench-v2~\citep{bai2024longbench2} and some Math tasks~\citep{ cobbe2021trainingverifierssolvemath, aime2024}}.
\small
\resizebox{\textwidth}{!}{
\begin{tabular}{l | c c c c c c | c | c c |c | c  c |c}
\toprule
\multirow{2.5}{*}{\textbf{Models}} & \multicolumn{7}{c|}{\textbf{\texttt{RULER}}} & \multicolumn{3}{c|}{\textbf{\texttt{LongBench-v2}}} & \multicolumn{3}{c}{\textbf{\texttt{Math}}} \\
\cmidrule(lr){2-8} \cmidrule(lr){9-11} \cmidrule(lr){12-14}
 & \bf 8K &\bf 16K &\bf 32K &\bf 64K &\bf 128K &\bf 256K & \textbf{Perf.} &\bf Easy &\bf Hard & \textbf{Perf.} &\bf GSM8K &\bf AIME24 & \textbf{Perf.} \\
\arrayrulecolor{black}\midrule
\rowcolor{blue!10}
\multicolumn{14}{c}{\textbf{Qwen3-4B backbone model}} \\
\arrayrulecolor{black!20}\midrule
Qwen3-4B & 87.49 & 86.82 & 60.05 & 70.98 & 53.19 & 43.27 & 66.00 & 32.67 & 22.18 & 25.96 & 39.70 & 30.35 & 35.03\\
+ DuoAttention & 79.38 & 76.08 & 52.91 & 69.02 & 43.28 & 44.96 & 60.67 & \textbf{31.33} & 24.06 & 26.68 & 39.70 & 37.05 & 38.38\\
+ PruLong & 74.21 & 75.72 & 47.88 & 59.27 & 47.10 & 45.69 & 60.25 & 28.00 & 25.56 & 26.44 & 39.70 & 30.35 & 35.03\\
+ TriangleMix & \textbf{87.42} & \textbf{85.10} & 58.73 & 67.94 & 50.97 & 44.47 & 63.74 & \textbf{31.33} & 22.18 & 25.48 & 40.30 & \underline{37.25} & 38.78\\
\arrayrulecolor{black!40}\midrule
+ FluxAttn (FA-SSA) & 81.58 & 82.11 & 58.73 & \textbf{72.89} & \underline{52.81} & \textbf{56.91} & \underline{66.95} & 29.33 & \textbf{28.57} & \textbf{28.85} & 40.30 & 37.05 & 38.68\\
+ FluxAttn (FA-XA) & \underline{86.79} & \underline{84.94} & \underline{59.52} & 68.82 & 51.77 & 43.43 & 63.67 & \underline{30.00} & 24.06 & 26.20 & \underline{42.20} & \textbf{40.35} & \underline{41.28}\\
+ FluxAttn (FA-TA) & 84.28 & 84.53 & \textbf{60.58} & 68.60 & 51.91 & 51.64 & 65.55 & \textbf{31.33} & 26.32 & \underline{28.12} & \textbf{45.00} & \textbf{40.35} & \textbf{42.68}\\
\rowcolor{black!15}
+ FluxAttn~(FA-SSA) & 80.36 & 80.75 & 56.08 & \underline{71.49} & \textbf{59.17} & \underline{56.00} & \textbf{67.19} & 28.00 & \underline{28.20} & \underline{28.12} & 39.90 & \underline{37.25} & 38.58\\

\arrayrulecolor{black}\midrule
\rowcolor{blue!25}
\multicolumn{14}{c}{\textbf{Qwen3-8B backbone model}} \\
\arrayrulecolor{black!20}\midrule
Qwen3-8B & 89.69 & 85.62 & 63.23 & 82.39 & 65.84 & 66.71 & 75.74 & 39.33 & 27.82 & 31.97 & 40.60 & 32.35 & 36.48\\
+ DuoAttention & \underline{86.68} & \underline{86.01} & 63.23 & 77.52 & 61.50 & 61.95 & 72.41 & \textbf{40.67} & 25.56 & 31.01 & 41.20 & 35.65 & 38.43\\
+ PruLong & 83.85 & 80.86 & 60.05 & 77.25 & 62.54 & 61.49 & 70.97 & 36.00 & 28.20 & 31.01 & 40.40 & 32.35 & 36.38\\
+ TriangleMix & 81.01 & 75.67 & \underline{63.49} & 73.76 & 61.54 & \textbf{66.84} & 70.47 & 36.00 & 27.44 & 30.53 & 41.20 & \textbf{44.15} & 42.68\\
\arrayrulecolor{black!40}\midrule
+ FluxAttn (FA-SSA) & 84.09 & 81.90 & 60.58 & \underline{79.30} & 64.74 & 65.27 & 73.03 & 36.67 & \underline{29.32} & 31.97 & \textbf{46.90} & 42.35 & \textbf{44.63}\\
+ FluxAttn (FA-XA) & 85.88 & 85.54 & \textbf{65.08} & \textbf{81.95} & \textbf{65.09} & \underline{65.38} & \textbf{74.65} & 32.67 & \textbf{32.71} & \textbf{32.69} & 43.20 & 35.65 & 39.43 \\
+ FluxAttn (FA-TA) & \textbf{87.49} & \textbf{86.17} & 60.85 & 78.72 & 60.75 & 63.03 & \underline{73.51} & 37.33 & 27.44 & 31.01 & 43.00 & 39.05 & 41.03\\
\rowcolor{black!15}
+ FluxAttn~(FA-SSA) & 83.54 & 81.00 & 59.79 & 77.93 & \underline{64.83} & 65.12 & 72.51 & \underline{39.33} & 28.20 & \underline{32.21} & \underline{45.30} & \underline{43.20} & \underline{44.25}\\

\arrayrulecolor{black}\midrule
\rowcolor{cyan!20}
\multicolumn{14}{c}{\textbf{Llama-3.1-8B-Instruct backbone model}} \\
\arrayrulecolor{black!20}\midrule
Llama-3.1-8B-Instruct & 92.88 & 92.83 & 89.46 & 70.79 & 80.12 & 72.34 & 83.47 & 32.00 & 33.08 & 32.69 & 42.30 & 30.35 & 36.33 \\
+ DuoAttention & 91.71 & 86.35 & 85.65 & 62.65 & 62.30 & 38.69 & 70.33 & 26.67 & 28.57 & 27.88 & 44.40 & 33.65 & 39.03 \\
+ PruLong & 86.96 & 76.55 & 70.65 & 54.52 & 48.18 & 30.00 & 59.87 & 30.00 & 24.44 & 26.44 & 41.30 & 29.85 & 35.58 \\
+ TriangleMix & \underline{92.44} & \underline{90.76} & \underline{86.75} & \underline{68.00} & \underline{78.25} & \underline{64.39} & \underline{80.46} & 29.33 & 25.56 & 26.92 & \underline{46.30} & 37.05 & \underline{41.68} \\
\arrayrulecolor{black!40}\midrule
+ FluxAttn (FA-SSA) & 82.88 & 78.09 & 70.39 & 52.29 & 62.20 & 50.73 & 76.75 & 34.00 & 28.95 & 30.77 & 45.30 & 37.05 & 41.18 \\
+ FluxAttn (FA-XA) & 92.43 & \textbf{90.85} & \textbf{88.23} & \textbf{68.56} & 75.86 & 60.80 & 79.51 & \textbf{36.00} & \textbf{31.95} & \textbf{33.41} & 44.40 & 33.65 & 39.03 \\
+ FluxAttn (FA-TA) & \textbf{92.72} & 90.53 & 86.45 & 67.78 & \textbf{80.63} & \textbf{67.09} & \textbf{81.50} & \underline{34.67} & 28.95 & 31.01 & \textbf{46.90} & \textbf{38.30} & \textbf{42.60} \\
\rowcolor{black!15}
+ FluxAttn~(FA-SSA) & 90.11 & 79.39 & 79.22 & 56.08 & 62.94 & 59.39 & 73.67 & \underline{34.67} & \underline{30.08} & \underline{31.73} & 45.90 & \underline{37.35} & 41.63 \\
\arrayrulecolor{black}\bottomrule
\end{tabular}
}
\vspace{-0.5em}
\label{tab:ruler_lb2_result}
\end{table*}

\subsection{Overall Inference Efficiency}
\label{exp:speedup}

\begin{figure}[t]
    \centering
    \begin{subfigure}{0.48\linewidth}
        \centering
        \includegraphics[width=\linewidth]{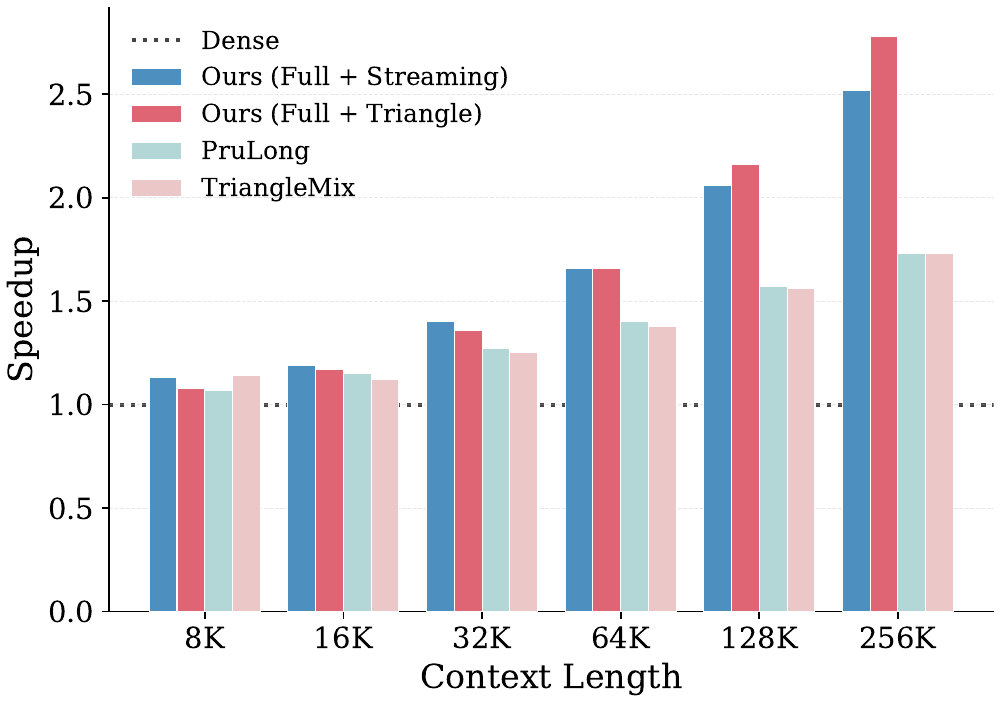}
        \caption{End-to-end speedup in the prefill phase.}
    \end{subfigure}
    \hfill
    \begin{subfigure}{0.48\linewidth}
        \centering
        \includegraphics[width=\linewidth]{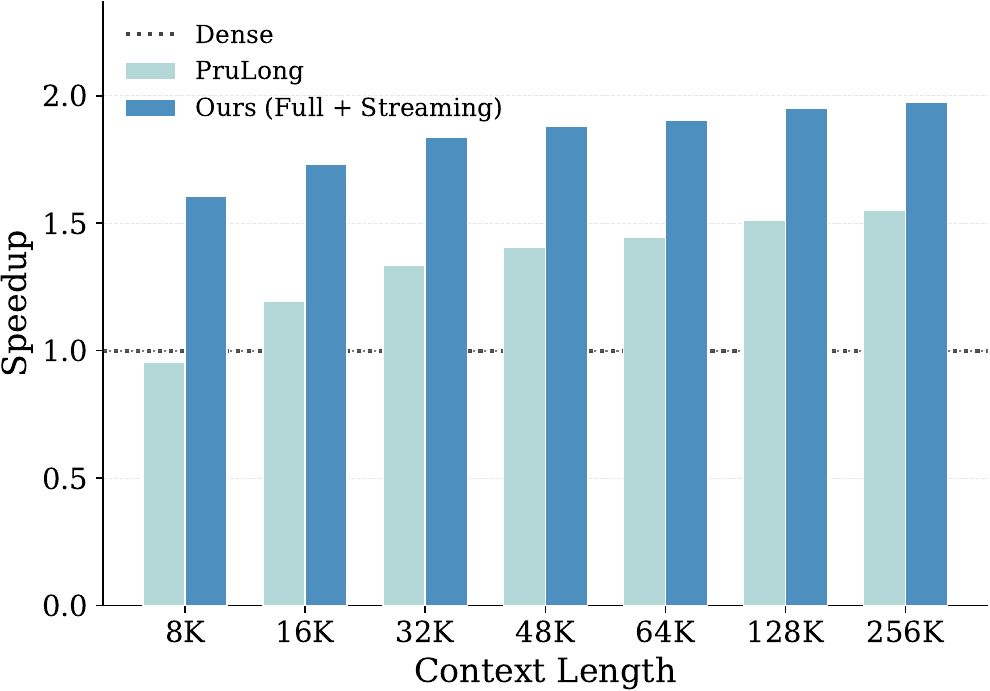}
        \caption{Kernel speedup in the decode phase.}
    \end{subfigure}
    \caption{Speedup comparison across different context lengths. The dotted line represents the dense baseline performance (1.0x).}
    \label{fig:speedup}
    \vspace{-1em}
\end{figure}

To evaluate the hardware acceleration of our method, we benchmark the inference speedup of FluxAttn against the standard dense baseline and existing sparse methods across varying context lengths. Figure~\ref{fig:speedup} presents the speedup metrics for both the prefill and decode phases.

\paragraph{End-to-End Prefill Acceleration}
Figure~\ref{fig:speedup}(a) shows the end-to-end latency reduction during the compute-bound prefill phase, i.e., time-to-first-token. As the context window expands, the quadratic complexity of standard attention becomes a bottleneck, allowing our dynamic routing mechanism to demonstrate substantial gains. At a 256K context length, our method (configured with FA-TA, i.e., full attention in retrieval layers and Triangle Attention in sparse layers) achieves up to a 2.8$\times$ end-to-end speedup, outperforming static baselines such as PruLong and TriangleMix.

\paragraph{Kernel-Level Decode Acceleration}
The advantage of layer-level routing is evident during the memory-bandwidth-bound decode phase, as shown in Figure~\ref{fig:speedup}(b). While prior approaches like PruLong struggle to translate theoretical sparsity into proportional wall-clock speedups due to fragmented memory access, FluxAttn solves this bottleneck by operating at the layer level. Our method achieves a scalable kernel speedup, approaching 2.0$\times$ at a 256K context length. This result empirically shows that our context-aware, layer-wise routing aligns with modern GPU execution patterns to deliver improved inference efficiency.
Note that the two panels report different phases and units---prefill latency and per-token decode latency---and must not be compared against each other; in particular, head-level kernels remain competitive on prefill, and our claim concerns per-token decode.

\paragraph{Quality--Latency--Memory Trade-off at Matched Sparsity}
The comparisons above vary the sparsity each method selects, which confounds routing granularity with the amount of computation retained. We therefore compare dense FA2, PruLong, and Flux Attention on Llama-3.1-8B-Instruct at approximately matched model sparsity ($\Omega_{\mathrm{MSR}} \approx 0.50$), jointly reporting accuracy, prefill and decode speedups, and persistent KV memory in a single table (Tables~\ref{tab:pareto} and~\ref{tab:pareto_full}, Appendix~\ref{appdix:pareto_full}). At 128K and 256K, where $\Omega_{\mathrm{MSR}}$ is exactly matched, Flux improves accuracy over PruLong by 14.76 and 29.39 points while reducing decode latency by 19.2\% and 32.0\%; persistent KV memory falls by 16.1--39.3\%, because a routed sparse layer need not retain the complete historical KV state whereas mixed-head execution does. Since a ``sparse unit'' is a head for PruLong and a layer for Flux, we report realized $\Omega_{\mathrm{MSR}}$ together with measured latency and memory rather than treating the ratio as a proxy for systems cost, and we note that this reflects the optimized head-wise implementation we evaluate rather than every possible head-wise kernel.

\paragraph{Batched Serving}
Single-request profiling can overstate the benefit of a sparsity scheme, because heterogeneous per-head workloads may be co-scheduled once the server batches requests. At a 32K context on $8\times$NVIDIA H20 GPUs, Flux stays faster at every feasible batch size (1.32$\times$ at batch 8, 2.07$\times$ at batch 16) and reaches batch 24 while FA2 exhausts memory (Appendix~\ref{appdix:batch_serving}).

\paragraph{Router Overhead Analysis}
A critical requirement for dynamic routing is minimizing its own computational cost. As illustrated in Figure~\ref{fig:router_latency}, our router incurs a negligible overhead, averaging only 0.20 ms per layer. Notably, the design exhibits length-invariant stability, maintaining a constant execution speed across sequence lengths ranging from 512 to 1M tokens. This ensures that the routing mechanism itself does not become a bottleneck at extreme context lengths, thereby preserving the substantial speedups achieved in the prefill phase.
\section{Analysis}
\label{sec:analysis}


\subsection{Dynamic Allocation Strategy of the Layer Router}
\label{ana:router_anlysis}

\begin{wrapfigure}{r}{0.45\linewidth}
  \centering
  \vspace{-1em}
        \includegraphics[width=\linewidth]{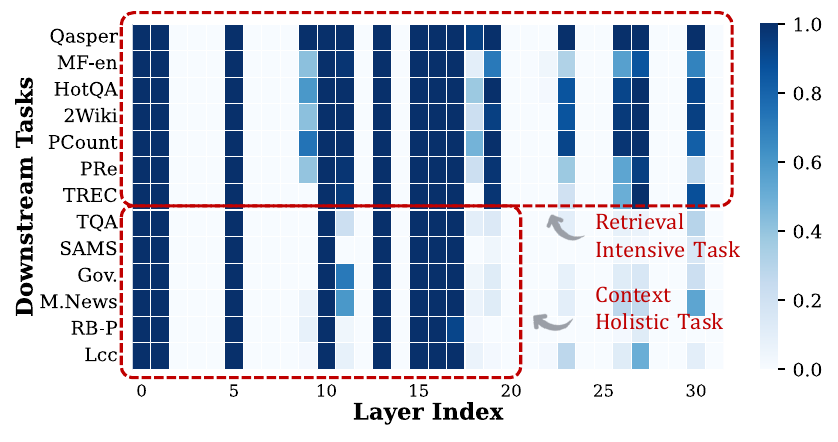}
    \caption{Overview of the layer-wise routing activation frequencies in Llama-3.1-8B-Instruct. Dark blue indicates layers consistently routed to FA across all six tasks in LongBench-E, whereas light blue denotes layers consistently routed to SA.}
    \label{fig:layer_activation}
    \vspace{-1em}
\end{wrapfigure}


\paragraph{Task-Level Dynamic Sparsity} 
Different downstream tasks impose inherently distinct requirements on attention sparsity. As shown in the upper region of Figure~\ref{fig:layer_activation}, retrieval-intensive tasks frequently activate FA (dark blue) to support the dense token interactions required for fact-finding. Conversely, context-holistic tasks predominantly route the mid-to-high layers to SA, which validates that high-level holistic semantic understanding is highly robust to attention sparsification. 
This demonstrates that Flux Attention replaces static allocations with task-aware dynamic sparsity.

\paragraph{Context-Aware Intra-Task Sparsity}
Beyond cross-task adaptation, the router further captures the intrinsic sparsity requirements of individual input contexts, rather than merely memorizing coarse-grained task-level patterns.
This instance-level variance is evident where intermediate activation frequencies ($\sim 0.4-0.6$, light blue) within a single task show the router adjusting to the complexity of different inputs.
We also find that specific layers (e.g., layers 0, 1, 5, 13, and 15--17) are consistently routed to FA across all tasks. This indicates the router preserves the universal architectural properties of the backbone while allocating the remaining computational budget based on specific task and context demands.

Notably, the emergence of this fine-grained, task-aware routing relies on a well-balanced training curriculum. An unbalanced data distribution can cause the router to collapse into a homogenized routing strategy, as extensively analyzed in Appendix \ref{app:data_composition}.
Furthermore, we find that a prefill-suffix pooling operation on the boundary 100 tokens is highly effective in driving this context-aware routing, as it isolates essential instruction signals from sequence noise (detailed in Appendix~\ref{appdix:truncation_analysis}).
\begin{figure*}[t]
    \centering
    \includegraphics[width=0.9\linewidth]{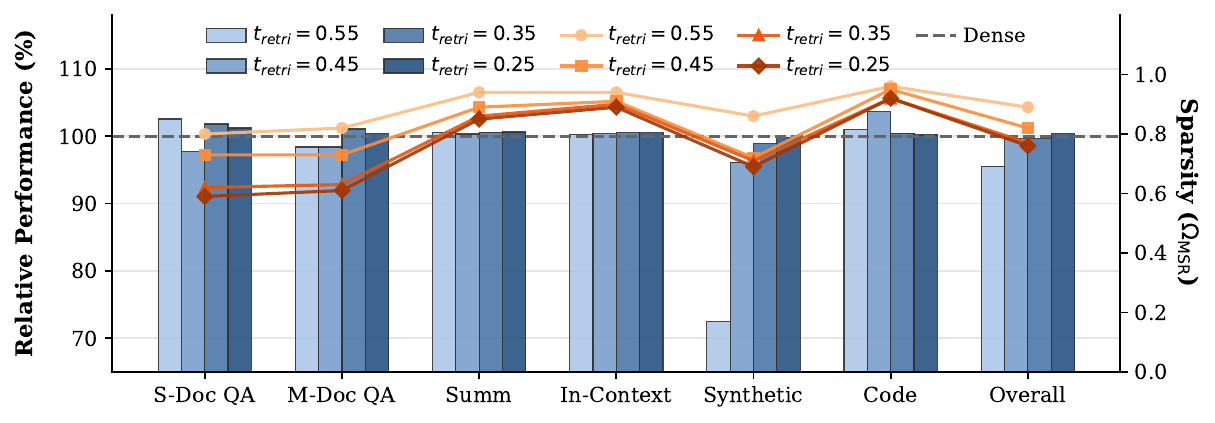}
    \caption{Comparison of performance and test-time $\Omega_{\mathrm{MSR}}$ among different training sparsity target $\boldsymbol{t}$ settings. The bar chart denotes the performance and the line chart denotes $\Omega_{\mathrm{MSR}}$ in each task.}
    \label{fig:diff_end_sparsity}
    \vspace{-1em}
\end{figure*}
\subsection{Impact of Target Sparsity Allocation}
\label{ana:sparsity_ratio_allocation}
We study the impact of target sparsity $\boldsymbol{t}$ on model performance. 
Specifically, we fix the target sparsity of context-holistic tasks to 1, while progressively decreasing target sparsity for retrieval-intensive tasks ($\boldsymbol{t}_\mathrm{retri}$) from 0.55 to 0.25.
Since $\boldsymbol{t}$ is a target for $\Omega_{\mathrm{MSR}}$ (Section~\ref{sec:method}), a lower $\boldsymbol{t}_\mathrm{retri}$ means \emph{less} Sparse Attention and \emph{more} Full Attention on retrieval-intensive tasks, and vice versa.
As shown in Figure~\ref{fig:diff_end_sparsity}, decreasing $\boldsymbol{t}_\mathrm{retri}$ causes the resulting ($\Omega_{\mathrm{MSR}}$) allocated by the model to exhibit slightly greater task-level differentiation across different tasks. 
However, $\Omega_{\mathrm{MSR}}$ does not strictly match the target $\boldsymbol{t}$.
This discrepancy arises because we use \emph{task-dependent and non-tight constraints}, which do not force the model to exactly satisfy the prescribed sparsity. 
We provide full training curves and further explanations in Appendix~\ref{appdix:loss_curve_monitoring_metrics}.
Additionally, when $\boldsymbol{t}_\mathrm{retri}$ is set too low (e.g., 0.25) to allocate a higher proportion of FA computation, the overall performance can even surpass that of the backbone model. 
Conversely, setting $\boldsymbol{t}_\mathrm{retri}$ too high causes the performance on retrieval-intensive tasks to drop sharply, consistent with the observations in Section~\ref{subsec:observations}.
To optimize inference efficiency, we adopt $\boldsymbol{t}_\mathrm{retri}=0.45$ in our main experiments, which achieves a favorable balance between strong overall performance and computational cost.

\subsection{Scalability via Backbone Adaptation}
\label{ana:sparsity_ada}

\begin{wrapfigure}{r}{0.45\linewidth}
  \centering
  \vspace{-1em}
        \includegraphics[width=\linewidth]{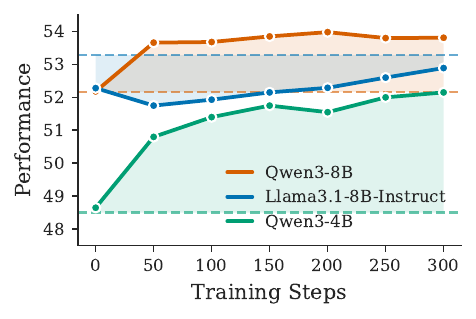}
    \caption{Performance trajectories during continued training with a frozen Layer Router. The backbone effectively adapts its representations to the established sparse pathways, demonstrating steady improvement over time.}
    \label{fig:continued_train}
\end{wrapfigure}

To evaluate the flexibility of Flux Attention, we investigate how well the method supports continued training. A critical question for dynamic sparsity methods is whether the routing mechanism can be decoupled from the backbone for subsequent model adaptation. To test this, we freeze the weights of the trained Layer Router, which fixes its learned dynamic allocation strategy, and continue training the model backbone using the data mixture from Section~\ref{subsec:exp_settings}.

As Figure~\ref{fig:continued_train} illustrates, continued training yields steady performance improvements across different models. Notably, both Qwen3-8B and Qwen3-4B rapidly surpass their original backbone performance (dashed lines) within just 50 steps and maintain a significant gain. While Llama3.1-8B-Instruct initially falls below its baseline, it demonstrates strong and continuous recovery throughout the training process, steadily closing the performance gap.
We attribute this delayed convergence to the heightened sensitivity of instruction-tuned models, which require additional steps to realign their complex representations under forced sparsity constraints.
These trends indicate that the backbone can effectively adapt its representations to the prescribed sparse pathways. Flux Attention thus offers practical post-training flexibility, allowing users to lock in an efficiency budget and fine-tune for downstream applications without disrupting the routing dynamics.

\section{Conclusion}
\label{sec:conclusion}

We introduce \textit{Flux Attention}, a context-aware dynamic routing framework mitigating the quadratic computational bottleneck of Large Language Models in long-context scenarios.
Unlike existing hybrid attention mechanisms relying on rigid static allocations or hardware-inefficient head-level routing, our approach employs a lightweight Layer Router adaptively assigning each transformer layer to full or sparse Attention based on task and input demands.
Extensive evaluations demonstrate our parameter-efficient method, requiring only 12 hours of training, achieves speedups up to 2.8$\times$ during prefilling and 2.0$\times$ during autoregressive decoding.
Crucially, it preserves high-fidelity information recovery across diverse long-context benchmarks, establishing a superior and scalable trade-off between generation quality and inference efficiency for modern LLMs.

\begin{ack}
We want to thank all the anonymous reviewers for their valuable comments. This work was supported by the National Science Foundation of China (NSFC No.~62576232) and sponsored by CCF-Baidu Open Fund.
\end{ack}

\small

\bibliographystyle{plain}
\bibliography{main}

@article{liu2025comprehensive,
  title={A comprehensive survey on long context language modeling},
  author={Liu, Jiaheng and Zhu, Dawei and Bai, Zhiqi and He, Yancheng and Liao, Huanxuan and Que, Haoran and Wang, Zekun and Zhang, Chenchen and Zhang, Ge and Zhang, Jiebin and others},
  journal={arXiv preprint arXiv:2503.17407},
  year={2025}
}

@misc{guo2024blocksparse,
  author       = {Guo, Junxian and Tang, Haotian and Yang, Shang and Zhang, Zhekai and Liu, Zhijian and Han, Song},
  title        = {{Block Sparse Attention}},
  year         = {2024},
  publisher    = {GitHub},
  journal      = {GitHub repository},
  howpublished = {\url{https://github.com/mit-han-lab/Block-Sparse-Attention}}
}

@article{child2019generating,
  title={Generating long sequences with sparse transformers},
  author={Child, Rewon},
  journal={arXiv preprint arXiv:1904.10509},
  year={2019}
}

@article{jang2016categorical,
  title={Categorical reparameterization with gumbel-softmax},
  author={Jang, Eric and Gu, Shixiang and Poole, Ben},
  journal={arXiv preprint arXiv:1611.01144},
  year={2016}
}

@article{peng2025accelerating,
  title={Accelerating Prefilling for Long-Context LLMs via Sparse Pattern Sharing},
  author={Peng, Dan and Fu, Zhihui and Ye, Zewen and Song, Zhuoran and Wang, Jun},
  journal={arXiv preprint arXiv:2505.19578},
  year={2025}
}

@article{jiang2024minference,
  title={Minference 1.0: Accelerating pre-filling for long-context llms via dynamic sparse attention},
  author={Jiang, Huiqiang and Li, Yucheng and Zhang, Chengruidong and Wu, Qianhui and Luo, Xufang and Ahn, Surin and Han, Zhenhua and Abdi, Amir H and Li, Dongsheng and Lin, Chin-Yew and others},
  journal={Advances in Neural Information Processing Systems},
  volume={37},
  pages={52481--52515},
  year={2024}
}

@inproceedings{streamingllm,
  title={Efficient Streaming Language Models with Attention Sinks},
  author={Guangxuan Xiao and Yuandong Tian and Beidi Chen and Song Han and Mike Lewis},
  booktitle={The Twelfth International Conference on Learning Representations},
  year={2024},
  url={https://openreview.net/forum?id=NG7sS51zVF}
}

@misc{TriangleMix,
      title={TriangleMix: Accelerating Prefilling via Decoding-time Contribution Sparsity}, 
      author={Zhiyuan He and Yike Zhang and Chengruidong Zhang and Huiqiang Jiang and Yuqing Yang and Lili Qiu},
      year={2025},
      eprint={2507.21526},
      archivePrefix={arXiv},
      primaryClass={cs.CL},
      url={https://arxiv.org/abs/2507.21526}, 
}

@article{Beltagy2020Longformer,
  title={Longformer: The Long-Document Transformer},
  author={Iz Beltagy and Matthew E. Peters and Arman Cohan},
  journal={arXiv:2004.05150},
  year={2020},
}

@article{li2025aixcoder,
  title={aiXcoder-7B-v2: Training LLMs to Fully Utilize the Long Context in Repository-level Code Completion},
  author={Li, Jia and Zhu, Hao and Liu, Huanyu and Shi, Xianjie and Zong, He and Dong, Yihong and Zhang, Kechi and Jiang, Siyuan and Jin, Zhi and Li, Ge},
  journal={arXiv preprint arXiv:2503.15301},
  year={2025}
}

@InProceedings{xsum-emnlp,
  author =      "Shashi Narayan and Shay B. Cohen and Mirella Lapata",
  title =       "Don't Give Me the Details, Just the Summary! {T}opic-Aware Convolutional Neural Networks for Extreme Summarization",
  booktitle =   "Proceedings of the 2018 Conference on Empirical Methods in Natural Language Processing ",
  year =        "2018",
  address =     "Brussels, Belgium",
}

@inproceedings{huang-etal-2021-efficient,
    title = "Efficient Attentions for Long Document Summarization",
    author = "Huang, Luyang  and
      Cao, Shuyang  and
      Parulian, Nikolaus  and
      Ji, Heng  and
      Wang, Lu",
    booktitle = "Proceedings of the 2021 Conference of the North American Chapter of the Association for Computational Linguistics: Human Language Technologies",
    month = jun,
    year = "2021",
    address = "Online",
    publisher = "Association for Computational Linguistics",
    url = "https://aclanthology.org/2021.naacl-main.112",
    doi = "10.18653/v1/2021.naacl-main.112",
    pages = "1419--1436",
}

@article{trivedi2022musique,
  title={MuSiQue: Multihop Questions via Single-hop Question Composition},
  author={Trivedi, Harsh and Balasubramanian, Niranjan and Khot, Tushar and Sabharwal, Ashish},
  journal={Transactions of the Association for Computational Linguistics},
  volume={10},
  pages={539--554},
  year={2022},
  publisher={MIT Press One Broadway, 12th Floor, Cambridge, Massachusetts 02142, USA~…}
}

@article{xu2024chatqa,
  title={ChatQA 2: Bridging the Gap to Proprietary LLMs in Long Context and RAG Capabilities},
  author={Xu, Peng and Ping, Wei and Wu, Xianchao and Liu, Zihan and Shoeybi, Mohammad and Catanzaro, Bryan},
  journal={arXiv preprint arXiv:2407.14482},
  year={2024}
}

@inproceedings{xiaoduoattention2025,
  title={DuoAttention: Efficient Long-Context LLM Inference with Retrieval and Streaming Heads},
  author={Xiao, Guangxuan and Tang, Jiaming and Zuo, Jingwei and Yang, Shang and Tang, Haotian and Fu, Yao and Han, Song and others},
  booktitle={The Thirteenth International Conference on Learning Representations},
  year={2025}
}

@article{zhang2025efficient,
  title={Efficient Context Scaling with LongCat ZigZag Attention},
  author={Zhang, Chen and Bai, Yang and Li, Jiahuan and Gui, Anchun and Wang, Keheng and Liu, Feifan and Wu, Guanyu and Jiang, Yuwei and Bu, Defei and Wei, Li and others},
  journal={arXiv preprint arXiv:2512.23966},
  year={2025}
}

@article{zaheer2020big,
  title={Big bird: Transformers for longer sequences},
  author={Zaheer, Manzil and Guruganesh, Guru and Dubey, Kumar Avinava and Ainslie, Joshua and Alberti, Chris and Ontanon, Santiago and Pham, Philip and Ravula, Anirudh and Wang, Qifan and Yang, Li and others},
  journal={Advances in neural information processing systems},
  volume={33},
  pages={17283--17297},
  year={2020}
}

@article{vaswani2017attention,
  title={Attention is all you need},
  author={Vaswani, Ashish and Shazeer, Noam and Parmar, Niki and Uszkoreit, Jakob and Jones, Llion and Gomez, Aidan N and Kaiser, {\L}ukasz and Polosukhin, Illia},
  journal={Advances in neural information processing systems},
  volume={30},
  year={2017}
}

@article{mei2025survey,
  title={A survey of context engineering for large language models},
  author={Mei, Lingrui and Yao, Jiayu and Ge, Yuyao and Wang, Yiwei and Bi, Baolong and Cai, Yujun and Liu, Jiazhi and Li, Mingyu and Li, Zhong-Zhi and Zhang, Duzhen and others},
  journal={arXiv preprint arXiv:2507.13334},
  year={2025}
}

@inproceedings{spargeattn,
  title={Spargeattn: Accurate sparse attention accelerating any model inference},
  author={Zhang, Jintao and Xiang, Chendong and Huang, Haofeng and Wei, Jia and Xi, Haocheng and Zhu, Jun and Chen, Jianfei},
  booktitle={International Conference on Machine Learning (ICML)},
  year={2025}
}

@inproceedings{
flexprefill,
title={FlexPrefill: A Context-Aware Sparse Attention Mechanism for Efficient Long-Sequence Inference},
author={Xunhao Lai and Jianqiao Lu and Yao Luo and Yiyuan Ma and Xun Zhou},
booktitle={The Thirteenth International Conference on Learning Representations},
year={2025},
url={https://openreview.net/forum?id=OfjIlbelrT}
}

@article{scissorhands,
  title={Scissorhands: Exploiting the persistence of importance hypothesis for llm kv cache compression at test time},
  author={Liu, Zichang and Desai, Aditya and Liao, Fangshuo and Wang, Weitao and Xie, Victor and Xu, Zhaozhuo and Kyrillidis, Anastasios and Shrivastava, Anshumali},
  journal={Advances in Neural Information Processing Systems},
  volume={36},
  year={2024}
}

@inproceedings{h2o,
  author       = {Zhenyu Zhang and
                  Ying Sheng and
                  Tianyi Zhou and
                  Tianlong Chen and
                  Lianmin Zheng and
                  Ruisi Cai and
                  Zhao Song and
                  Yuandong Tian and
                  Christopher R{\'{e}} and
                  Clark W. Barrett and
                  Zhangyang Wang and
                  Beidi Chen},
  editor       = {Alice Oh and
                  Tristan Naumann and
                  Amir Globerson and
                  Kate Saenko and
                  Moritz Hardt and
                  Sergey Levine},
  title        = {{H2O:} Heavy-Hitter Oracle for Efficient Generative Inference of Large
                  Language Models},
  booktitle    = {Advances in Neural Information Processing Systems 36: Annual Conference
                  on Neural Information Processing Systems 2023, NeurIPS 2023, New Orleans,
                  LA, USA, December 10 - 16, 2023},
  year         = {2023},
  bibsource    = {dblp computer science bibliography, https://dblp.org}
}

@article{snapkv,
  title={Snapkv: Llm knows what you are looking for before generation},
  author={Li, Yuhong and Huang, Yingbing and Yang, Bowen and Venkitesh, Bharat and Locatelli, Acyr and Ye, Hanchen and Cai, Tianle and Lewis, Patrick and Chen, Deming},
  journal={arXiv preprint arXiv:2404.14469},
  year={2024}
}

@misc{peng2025acceleratingprefillinglongcontextllms,
      title={Accelerating Prefilling for Long-Context LLMs via Sparse Pattern Sharing}, 
      author={Dan Peng and Zhihui Fu and Zewen Ye and Zhuoran Song and Jun Wang},
      year={2025},
      eprint={2505.19578},
      archivePrefix={arXiv},
      primaryClass={cs.LG},
      url={https://arxiv.org/abs/2505.19578}, 
}

@misc{ji2025salelowbitestimation,
      title={SALE : Low-bit Estimation for Efficient Sparse Attention in Long-context LLM Prefilling}, 
      author={Xiaodong Ji and Hailin Zhang and Fangcheng Fu and Bin Cui},
      year={2025},
      eprint={2505.24179},
      archivePrefix={arXiv},
      primaryClass={cs.LG},
      url={https://arxiv.org/abs/2505.24179}, 
}

@article{wu2024retrieval,
  title={Retrieval head mechanistically explains long-context factuality},
  author={Wu, Wenhao and Wang, Yizhong and Xiao, Guangxuan and Peng, Hao and Fu, Yao},
  journal={arXiv preprint arXiv:2404.15574},
  year={2024}
}

@article{bhaskar2025cache,
  title={Cache Me If You Can: How Many KVs Do You Need for Effective Long-Context LMs?},
  author={Bhaskar, Adithya and Wettig, Alexander and Gao, Tianyu and Dong, Yihe and Chen, Danqi},
  journal={arXiv preprint arXiv:2506.17121},
  year={2025}
}

@article{grattafiori2024llama,
  title={The llama 3 herd of models},
  author={Grattafiori, Aaron and Dubey, Abhimanyu and Jauhri, Abhinav and Pandey, Abhinav and Kadian, Abhishek and Al-Dahle, Ahmad and Letman, Aiesha and Mathur, Akhil and Schelten, Alan and Vaughan, Alex and others},
  journal={arXiv preprint arXiv:2407.21783},
  year={2024}
}

@misc{yang2025qwen3technicalreport,
      title={Qwen3 Technical Report}, 
      author={An Yang and Anfeng Li and Baosong Yang and Beichen Zhang and Binyuan Hui and Bo Zheng and Bowen Yu and Chang Gao and Chengen Huang and Chenxu Lv and Chujie Zheng and Dayiheng Liu and Fan Zhou and Fei Huang and Feng Hu and Hao Ge and Haoran Wei and Huan Lin and Jialong Tang and Jian Yang and Jianhong Tu and Jianwei Zhang and Jianxin Yang and Jiaxi Yang and Jing Zhou and Jingren Zhou and Junyang Lin and Kai Dang and Keqin Bao and Kexin Yang and Le Yu and Lianghao Deng and Mei Li and Mingfeng Xue and Mingze Li and Pei Zhang and Peng Wang and Qin Zhu and Rui Men and Ruize Gao and Shixuan Liu and Shuang Luo and Tianhao Li and Tianyi Tang and Wenbiao Yin and Xingzhang Ren and Xinyu Wang and Xinyu Zhang and Xuancheng Ren and Yang Fan and Yang Su and Yichang Zhang and Yinger Zhang and Yu Wan and Yuqiong Liu and Zekun Wang and Zeyu Cui and Zhenru Zhang and Zhipeng Zhou and Zihan Qiu},
      year={2025},
      eprint={2505.09388},
      archivePrefix={arXiv},
      primaryClass={cs.CL},
      url={https://arxiv.org/abs/2505.09388}, 
}

@misc{loshchilov2019decoupledweightdecayregularization,
      title={Decoupled Weight Decay Regularization}, 
      author={Ilya Loshchilov and Frank Hutter},
      year={2019},
      eprint={1711.05101},
      archivePrefix={arXiv},
      primaryClass={cs.LG},
      url={https://arxiv.org/abs/1711.05101}, 
}

@inproceedings{bai2024longbench,
    title = "{L}ong{B}ench: A Bilingual, Multitask Benchmark for Long Context Understanding",
    author = "Bai, Yushi and Lv, Xin  and Zhang, Jiajie  and Lyu, Hongchang  and
      Tang, Jiankai  and Huang, Zhidian  and Du, Zhengxiao  and Liu, Xiao  and Zeng, Aohan  and Hou, Lei  and Dong, Yuxiao  and Tang, Jie  and Li, Juanzi",
    booktitle = "Proceedings of the 62nd Annual Meeting of the Association for Computational Linguistics (Volume 1: Long Papers)",
    month = aug,
    year = "2024",
    address = "Bangkok, Thailand",
    publisher = "Association for Computational Linguistics",
    url = "https://aclanthology.org/2024.acl-long.172",
    doi = "10.18653/v1/2024.acl-long.172",
    pages = "3119--3137",
}

@inproceedings{NSA,
    title = "Native Sparse Attention: Hardware-Aligned and Natively Trainable Sparse Attention",
    author = "Yuan, Jingyang  and
      Gao, Huazuo  and
      Dai, Damai  and
      Luo, Junyu  and
      Zhao, Liang  and
      Zhang, Zhengyan  and
      Xie, Zhenda  and
      Wei, Yuxing  and
      Wang, Lean  and
      Xiao, Zhiping  and
      Wang, Yuqing  and
      Ruan, Chong  and
      Zhang, Ming  and
      Liang, Wenfeng  and
      Zeng, Wangding",
    editor = "Che, Wanxiang  and
      Nabende, Joyce  and
      Shutova, Ekaterina  and
      Pilehvar, Mohammad Taher",
    booktitle = "Proceedings of the 63rd Annual Meeting of the Association for Computational Linguistics (Volume 1: Long Papers)",
    month = jul,
    year = "2025",
    address = "Vienna, Austria",
    publisher = "Association for Computational Linguistics",
    url = "https://aclanthology.org/2025.acl-long.1126/",
    doi = "10.18653/v1/2025.acl-long.1126",
    pages = "23078--23097",
    ISBN = "979-8-89176-251-0",
}

@misc{infllmv2,
      title={InfLLM-V2: Dense-Sparse Switchable Attention for Seamless Short-to-Long Adaptation}, 
      author={Weilin Zhao and Zihan Zhou and Zhou Su and Chaojun Xiao and Yuxuan Li and Yanghao Li and Yudi Zhang and Weilun Zhao and Zhen Li and Yuxiang Huang and Ao Sun and Xu Han and Zhiyuan Liu},
      year={2025},
      eprint={2509.24663},
      archivePrefix={arXiv},
      primaryClass={cs.CL},
      url={https://arxiv.org/abs/2509.24663}, 
}

@misc{moba,
      title={MoBA: Mixture of Block Attention for Long-Context LLMs}, 
      author={Enzhe Lu and Zhejun Jiang and Jingyuan Liu and Yulun Du and Tao Jiang and Chao Hong and Shaowei Liu and Weiran He and Enming Yuan and Yuzhi Wang and Zhiqi Huang and Huan Yuan and Suting Xu and Xinran Xu and Guokun Lai and Yanru Chen and Huabin Zheng and Junjie Yan and Jianlin Su and Yuxin Wu and Neo Y. Zhang and Zhilin Yang and Xinyu Zhou and Mingxing Zhang and Jiezhong Qiu},
      year={2025},
      eprint={2502.13189},
      archivePrefix={arXiv},
      primaryClass={cs.LG},
      url={https://arxiv.org/abs/2502.13189}, 
}

@article{duoattention,
        title={DuoAttention: Efficient Long-Context LLM Inference with Retrieval and Streaming Heads},
        author={Xiao, Guangxuan and Tang, Jiaming and Zuo, Jingwei and Guo, Junxian and Yang, Shang and Tang, Haotian and Fu, Yao and Han, Song},
        journal={arXiv},
        year={2024}
}

@inproceedings{xattention,
  title     = {XAttention: Block Sparse Attention with Antidiagonal Scoring},
  author    = {Xu, Ruyi and Xiao, Guangxuan and Huang, Haofeng and Guo, Junxian and Han, Song},
  booktitle = {Proceedings of the 42nd International Conference on Machine Learning (ICML)},
  year      = {2025}
}

@article{hsieh2024ruler,
  title={RULER: What's the Real Context Size of Your Long-Context Language Models?},
  author={Cheng-Ping Hsieh and Simeng Sun and Samuel Kriman and Shantanu Acharya and Dima Rekesh and Fei Jia and Yang Zhang and Boris Ginsburg},
  year={2024},
  journal={arXiv preprint arXiv:2404.06654},
}

@inproceedings{bai2024longbench2,
  title={Longbench v2: Towards deeper understanding and reasoning on realistic long-context multitasks},
  author={Bai, Yushi and Tu, Shangqing and Zhang, Jiajie and Peng, Hao and Wang, Xiaozhi and Lv, Xin and Cao, Shulin and Xu, Jiazheng and Hou, Lei and Dong, Yuxiao and others},
  booktitle={Proceedings of the 63rd Annual Meeting of the Association for Computational Linguistics (Volume 1: Long Papers)},
  pages={3639--3664},
  year={2025}
}

@article{tang2025loom,
    title={LOOM-Scope: a comprehensive and efficient LOng-cOntext Model evaluation framework},
    author={Tang, Zecheng and Wang, Haitian and Qiu, Quantong and Ji, Baibei and Sun, Ruoxi and Zhou, Keyan and Li, Juntao and Zhang, Min},
    journal={arXiv preprint arXiv:2507.04723},
    year={2025}
    }

@article{gao2024seerattention,
    title={SeerAttention: Learning Intrinsic Sparse Attention in Your LLMs},
    author={Gao, Yizhao and Zeng, Zhichen and Du, Dayou and Cao, Shijie and So, Hayden Kwok-Hay and Cao, Ting and Yang, Fan and Yang, Mao},
    journal={arXiv preprint arXiv:2410.13276},
    year={2024}
}

@misc{deepseekai2024deepseekv32,
      title={DeepSeek-V3.2-Exp: Boosting Long-Context Efficiency with DeepSeek Sparse Attention}, 
      author={DeepSeek-AI},
      year={2025},
}

@misc{lieber2024jambahybridtransformermambalanguage,
      title={Jamba: A Hybrid Transformer-Mamba Language Model}, 
      author={Opher Lieber and Barak Lenz and Hofit Bata and Gal Cohen and Jhonathan Osin and Itay Dalmedigos and Erez Safahi and Shaked Meirom and Yonatan Belinkov and Shai Shalev-Shwartz and Omri Abend and Raz Alon and Tomer Asida and Amir Bergman and Roman Glozman and Michael Gokhman and Avashalom Manevich and Nir Ratner and Noam Rozen and Erez Shwartz and Mor Zusman and Yoav Shoham},
      year={2024},
      eprint={2403.19887},
      archivePrefix={arXiv},
      primaryClass={cs.CL},
      url={https://arxiv.org/abs/2403.19887}, 
}

@article{ren2024samba,
      title={Samba: Simple Hybrid State Space Models for Efficient Unlimited Context Language Modeling}, 
      author={Liliang Ren and Yang Liu and Yadong Lu and Yelong Shen and Chen Liang and Weizhu Chen},
      journal = {arXiv preprint},
      year={2024},
      url={https://arxiv.org/abs/2406.07522}
}

@misc{glorioso2024zambacompact7bssm,
      title={Zamba: A Compact 7B SSM Hybrid Model}, 
      author={Paolo Glorioso and Quentin Anthony and Yury Tokpanov and James Whittington and Jonathan Pilault and Adam Ibrahim and Beren Millidge},
      year={2024},
      eprint={2405.16712},
      archivePrefix={arXiv},
      primaryClass={cs.LG},
      url={https://arxiv.org/abs/2405.16712}, 
}

@misc{ku2025systemsalgorithmsconvolutionalmultihybrid,
      title={Systems and Algorithms for Convolutional Multi-Hybrid Language Models at Scale}, 
      author={Jerome Ku and Eric Nguyen and David W. Romero and Garyk Brixi and Brandon Yang and Anton Vorontsov and Ali Taghibakhshi and Amy X. Lu and Dave P. Burke and Greg Brockman and Stefano Massaroli and Christopher Ré and Patrick D. Hsu and Brian L. Hie and Stefano Ermon and Michael Poli},
      year={2025},
      eprint={2503.01868},
      archivePrefix={arXiv},
      primaryClass={cs.LG},
      url={https://arxiv.org/abs/2503.01868}, 
}

@misc{dao2024transformersssmsgeneralizedmodels,
      title={Transformers are SSMs: Generalized Models and Efficient Algorithms Through Structured State Space Duality}, 
      author={Tri Dao and Albert Gu},
      year={2024},
      eprint={2405.21060},
      archivePrefix={arXiv},
      primaryClass={cs.LG},
      url={https://arxiv.org/abs/2405.21060}, 
}

@inproceedings{UnComp,
    title = "{UNC}omp: Can Matrix Entropy Uncover Sparsity? {---} A Compressor Design from an Uncertainty-Aware Perspective",
    author = "Xiong, Jing  and
      Shen, Jianghan  and
      Ye, Fanghua  and
      Tao, Chaofan  and
      Wan, Zhongwei  and
      Lu, Jianqiao  and
      Wu, Xun  and
      Zheng, Chuanyang  and
      Guo, Zhijiang  and
      Yang, Min  and
      Kong, Lingpeng  and
      Wong, Ngai",
    editor = "Christodoulopoulos, Christos  and
      Chakraborty, Tanmoy  and
      Rose, Carolyn  and
      Peng, Violet",
    booktitle = "Proceedings of the 2025 Conference on Empirical Methods in Natural Language Processing",
    month = nov,
    year = "2025",
    address = "Suzhou, China",
    publisher = "Association for Computational Linguistics",
    url = "https://aclanthology.org/2025.emnlp-main.209/",
    doi = "10.18653/v1/2025.emnlp-main.209",
    pages = "4179--4199",
    ISBN = "979-8-89176-332-6"
}

@inproceedings{
lin2026lycheedecode,
title={LycheeDecode: Accelerating Long-Context {LLM} Inference via Hybrid-Head Sparse Decoding},
author={Gang Lin and Dongfang Li and Zhuoen Chen and Yukun Shi and Xuhui Chen and Baotian Hu and Min Zhang},
booktitle={The Fourteenth International Conference on Learning Representations},
year={2026},
url={https://openreview.net/forum?id=YWCHLdNGVU}
}

@misc{tang2026elasticattentiontesttimeadaptive,
      title={Elastic Attention: Test-time Adaptive Sparsity Ratios for Efficient Transformers}, 
      author={Zecheng Tang and Quantong Qiu and Yi Yang and Zhiyi Hong and Haiya Xiang and Kebin Liu and Qingqing Dang and Juntao Li and Min Zhang},
      year={2026},
      eprint={2601.17367},
      archivePrefix={arXiv},
      primaryClass={cs.CL},
      url={https://arxiv.org/abs/2601.17367}, 
}

@article{aime2024,
  title={American Invitational Mathematics Examination (AIME)},
  author={MAA},
  series={Mathematics Competition Series,n.d.b},
  journal={URL https://maa.org/math-competitions/aime},
  year={2024}
}

@misc{cobbe2021trainingverifierssolvemath,
      title={Training Verifiers to Solve Math Word Problems}, 
      author={Karl Cobbe and Vineet Kosaraju and Mohammad Bavarian and Mark Chen and Heewoo Jun and Lukasz Kaiser and Matthias Plappert and Jerry Tworek and Jacob Hilton and Reiichiro Nakano and Christopher Hesse and John Schulman},
      year={2021},
      eprint={2110.14168},
      archivePrefix={arXiv},
      primaryClass={cs.LG},
      url={https://arxiv.org/abs/2110.14168}, 
}

@misc{raposo2024mixtureofdepthsdynamicallyallocatingcompute,
      title={Mixture-of-Depths: Dynamically allocating compute in transformer-based language models}, 
      author={David Raposo and Sam Ritter and Blake Richards and Timothy Lillicrap and Peter Conway Humphreys and Adam Santoro},
      year={2024},
      eprint={2404.02258},
      archivePrefix={arXiv},
      primaryClass={cs.LG},
      url={https://arxiv.org/abs/2404.02258}, 
}

@inproceedings{
shazeer2017,
title={ Outrageously Large Neural Networks: The Sparsely-Gated Mixture-of-Experts Layer},
author={Noam Shazeer and *Azalia Mirhoseini and *Krzysztof Maziarz and Andy Davis and Quoc Le and Geoffrey Hinton and Jeff Dean},
booktitle={International Conference on Learning Representations},
year={2017},
url={https://openreview.net/forum?id=B1ckMDqlg}
}

@misc{fedus2022switchtransformersscalingtrillion,
      title={Switch Transformers: Scaling to Trillion Parameter Models with Simple and Efficient Sparsity}, 
      author={William Fedus and Barret Zoph and Noam Shazeer},
      year={2022},
      eprint={2101.03961},
      archivePrefix={arXiv},
      primaryClass={cs.LG},
      url={https://arxiv.org/abs/2101.03961}, 
}

@article{zhang2024lighttransfer,
  title={Lighttransfer: Your long-context LLM is secretly a hybrid model with effortless adaptation},
  author={Zhang, Xuan and Zhang, Fengzhuo and Du, Cunxiao and Du, Chao and Pang, Tianyu and Gao, Wei and Lin, Min},
  journal={arXiv preprint arXiv:2410.13846},
  year={2024}
}

@inproceedings{desai2026vattention,
  title={vattention: Verified sparse attention via sampling},
  author={Desai, Aditya and Agrawal, Kumar and Yang, Shuo and Cuadron Lafuente, Alejandro and Gaspar Schroeder, Luis and Zaharia, Matei and Gonzalez, Joseph E and Stoica, Ion},
  booktitle={International Conference on Learning Representations},
  volume={2026},
  pages={52253--52284},
  year={2026}
}

@article{ren2023sparsemodular,
  title={Sparse modular activation for efficient sequence modeling},
  author={Ren, Liliang and Liu, Yang and Wang, Shuohang and Xu, Yichong and Zhu, Chenguang and Zhai, Cheng Xiang},
  journal={Advances in Neural Information Processing Systems},
  volume={36},
  pages={19799--19822},
  year={2023}
}

@article{zhang2026sla2,
  title={Sla2: Sparse-linear attention with learnable routing and qat},
  author={Zhang, Jintao and Wang, Haoxu and Jiang, Kai and Zheng, Kaiwen and Jiang, Youhe and Stoica, Ion and Chen, Jianfei and Zhu, Jun and Gonzalez, Joseph E},
  journal={arXiv preprint arXiv:2602.12675},
  year={2026}
}

@article{lancucki2025inferencetime,
  title={Inference-time hyper-scaling with kv cache compression},
  author={{\L}a{\'n}cucki, Adrian and Staniszewski, Konrad and Nawrot, Piotr and Ponti, Edoardo Maria},
  journal={Advances in Neural Information Processing Systems},
  volume={38},
  pages={9365--9397},
  year={2026}
}

@article{le2026sketchandwalk,
  title={Scout before you attend: Sketch-and-walk sparse attention for efficient LLM inference},
  author={Le, Hoang Anh Duy and Joshi, Sahil and Yang, Zeyu and Xu, Zhaozhuo and Shrivastava, Anshumali},
  journal={arXiv preprint arXiv:2602.07397},
  year={2026}
}

@article{wang2026pbsattn,
  title={Sparser Block-Sparse Attention via Token Permutation},
  author={Wang, Xinghao and Wang, Pengyu and Zhang, Dong and Tan, Chenkun and Zhou, Shaojun and Liu, Zhaoxiang and Lian, Shiguo and Liu, Fangxu and Song, Kai and Qiu, Xipeng},
  journal={arXiv preprint arXiv:2510.21270},
  year={2025}
}

@inproceedings{quest,
  author       = {Jiaming Tang and
                  Yilong Zhao and
                  Kan Zhu and
                  Guangxuan Xiao and
                  Baris Kasikci and
                  Song Han},
  editor       = {Ruslan Salakhutdinov and
                  Zico Kolter and
                  Katherine A. Heller and
                  Adrian Weller and
                  Nuria Oliver and
                  Jonathan Scarlett and
                  Felix Berkenkamp},
  title        = {{QUEST:} Query-Aware Sparsity for Efficient Long-Context {LLM} Inference},
  booktitle    = {Forty-first International Conference on Machine Learning, {ICML} 2024,
                  Vienna, Austria, July 21-27, 2024},
  series       = {Proceedings of Machine Learning Research},
  volume       = {235},
  pages        = {47901--47911},
  publisher    = {{PMLR} / OpenReview.net},
  year         = {2024},
  url          = {https://proceedings.mlr.press/v235/tang24l.html},
  bibsource    = {dblp computer science bibliography, https://dblp.org}
}

\clearpage


\appendix

\section{Code Release and Supplementary Material}
\label{appdix:anonymous_code}
Our code is available at \url{https://github.com/qqtang-code/FluxAttention}.

\section{Related Work}
\label{sec:related_work}

\subsection{Sparse and Selective Attention}
\label{subsec:sparse_attn}
To mitigate the quadratic complexity of standard attention mechanisms, existing research has broadly followed two main directions: inference-time heuristics and training-aware sparsification. Inference-time heuristics typically employ static patterns, such as fixed sliding windows or strides~\citep{streamingllm, TriangleMix, Beltagy2020Longformer}, to restrict the receptive field. To capture dynamic dependencies more effectively, content-aware approaches discard uninformative tokens based on accumulated importance scores~\citep{h2o, snapkv, scissorhands}, or use kernel-based estimators to identify important blocks and bypass redundant computations~\citep{jiang2024minference}. Additionally, prefill optimizers apply importance-driven selection to accelerate the processing of long contexts~\citep{flexprefill, xattention, spargeattn, peng2025acceleratingprefillinglongcontextllms, ji2025salelowbitestimation}. Recent block-level methods refine this further: Sketch-and-Walk~\citep{le2026sketchandwalk} uses lightweight sketches and cross-layer walk scores to select top-$k$ blocks for both prefill and decode, while PBS-Attn~\citep{wang2026pbsattn} permutes tokens to expose greater block sparsity and accelerates prefill with customized kernels. Verified sparsity takes a different route: vAttention~\citep{desai2026vattention} samples attention weights to certify that a per-head, per-token sparsity pattern preserves the full-attention output. Despite their effectiveness, these heuristic methods frequently rely on sensitive hyperparameters, limiting their robustness across different tasks.

In contrast, training-aware sparsification incorporates sparsity into the optimization objective to align the training process with sparse inference. A prominent direction in this area involves learnable selection. For instance, SeerAttention~\citep{gao2024seerattention}, NSA~\citep{NSA}, and MoBA~\citep{moba} employ learnable gates and hierarchical constraints to approximate ground-truth attention patterns. To connect dense pre-training with sparse adaptation, InfLLM-v2~\citep{infllmv2} introduces a dense-sparse switchable mechanism via parameter-free pooling, whereas DSA~\citep{deepseekai2024deepseekv32} utilizes a lightning indexer alongside a two-stage training strategy to efficiently filter the top-$k$ key-value pairs. However, most of these methods focus on fine-grained block-level or token-level selection within a fixed attention framework, rather than dynamically adapting the overall attention mode based on input complexity.

The operators above determine \emph{which tokens or blocks} are attended. Flux Attention is orthogonal: it does not introduce a new sparse operator, and the results in this paper cover three materially different backends---SSA, XAttention, and Triangle Attention. Any of the block-level or fine-grained operators above could in principle serve as the sparse branch of a routed layer; we describe this as architectural compatibility rather than measured integration, and we restrict performance claims to the backends we evaluate.

\subsection{Hybrid Architectures and Dynamic Allocation}
\label{subsec:hybrid_dynamic}
To balance computational efficiency and model performance, hybrid architectures strategically integrate Full Attention (FA) with linear-complexity operators. 
The dominant paradigm, inter-layer hybridization, interleaves linear layers with standard attention layers to recover associative recall capabilities~\citep{ku2025systemsalgorithmsconvolutionalmultihybrid, dao2024transformersssmsgeneralizedmodels}. Notable large-scale implementations, such as Jamba~\citep{lieber2024jambahybridtransformermambalanguage}, utilize fixed block-wise ratios, whereas variants optimize memory utilization through shared global blocks~\citep{glorioso2024zambacompact7bssm} or sliding windows~\citep{ren2024samba}.
Extending this paradigm to adapt existing dense models, LightTransfer~\citep{zhang2024lighttransfer} identifies specific ``lazy layers'' that primarily focus on recent or initial tokens, statically replacing their full attention with streaming attention to convert long-context LLMs into hybrid variants with effortless adaptation.
More recently, intra-layer hybridization has emerged as a strategy to refine structural granularity. For example, PruLong~\citep{bhaskar2025cache} and DuoAttention~\citep{duoattention} combine FA and Sparse Attention (SA) within individual layers by assigning different attention heads to different computational modes. Furthermore, LongCat~\citep{zhang2025efficient} proposes the LoZA mechanism, constructing a static ZigZag topology by replacing low-sensitivity Multi-head Latent Attention (MLA) modules with linear-complexity SA.
A critical limitation of these approaches is their reliance on static topologies or pre-defined ratios established prior to inference, lacking the flexibility required to dynamically distinguish diverse tasks.

Dynamic allocation does not make a method novel by itself. SeqBoat~\citep{ren2023sparsemodular} learns to activate a sparse attention module inside an otherwise recurrent model; Elastic Attention~\citep{tang2026elasticattentiontesttimeadaptive} allocates varying sparsity at the head level based on contextual importance; and SLA2~\citep{zhang2026sla2} learns a router that chooses between sparse and linear attention. Decode-time cache management is a parallel line of work, including inference-time hyper-scaling, which trades KV cache capacity for a longer generation budget~\citep{lancucki2025inferencetime}. What distinguishes Flux Attention is the \emph{granularity and the deployment objective}: it conditions a whole-layer FA/SA decision on the input, keeps the decision homogeneous inside each layer so that the savings survive into measured decode latency and KV memory, and evaluates that quality--decode-latency--memory trade-off rather than FLOP reduction alone. We therefore do not claim to be the first dynamic or learned sparse-attention method; Table~\ref{tab:related_work_comparison} positions our contribution along the unit of selection, training, inference phase, and objective.

One apparent conflict in the literature is worth resolving explicitly. Head-level dynamic sparsity has been reported to deliver substantial speedups, whereas our measurements show only marginal gains for it. The two results concern different inference phases: reported head-level fused-kernel gains are prefill/TTFT accelerations, while our Figure~\ref{fig:speedup}(b) and Appendix~\ref{appendix:latency_measurement} measure per-token autoregressive decode latency, where historical-KV access and memory bandwidth dominate. For decode, we additionally compare against PruLong's optimized block-sparse decode path in Table~\ref{tab:pareto}. Specialized head-wise kernels can therefore accelerate prefill, while the evaluated optimized head-wise implementation provides a smaller and less scalable decode benefit than whole-layer routing.

To address the rigidity of static designs, recent studies have explored dynamic allocation strategies. For instance, Elastic Attention~\citep{tang2026elasticattentiontesttimeadaptive} dynamically allocates varying sparsity at the head level based on contextual importance. While offering algorithmic flexibility, such head-level dynamic sparsity introduces hardware inefficiencies in the memory-bandwidth-bound decode phase. Specifically, varying context lengths across different attention heads lead to synchronization bottlenecks, as fast-executing sparse heads must wait for memory-intensive retrieval heads within the same layer. This creates significant memory bandwidth bottlenecks, hindering hardware acceleration and limiting practical speedups, especially during the autoregressive decoding phase.

\subsection{Dynamic Routing in Neural Networks}
\label{subsec:dynamic_routing}
Dynamic routing and conditional computation have been studied to decouple model capacity from inference cost. Traditional approaches, such as Mixture-of-Experts (MoE)~\citep{shazeer2017, fedus2022switchtransformersscalingtrillion}, route tokens to specialized Feed-Forward Network (FFN) experts. Recent methods like Mixture-of-Depths (MoD)~\citep{raposo2024mixtureofdepthsdynamicallyallocatingcompute} apply this idea by dynamically skipping specific layers for uninformative tokens to optimize compute allocation.
While these methods successfully route computation, they primarily focus on FFNs or complete layer skipping, leaving the dynamic optimization of the attention mechanism less explored. Flux Attention instead routes the attention \emph{mode} of an entire layer, which keeps intra-layer work homogeneous and memory access contiguous so that the theoretical reduction in attention compute is realized as wall-clock speedup.

\begin{table*}[t]
\centering
\caption{Positioning of Flux Attention against representative dynamic and fine-grained attention-selection methods. The axes are the unit at which computation is selected, whether the selection is learned or fixed, the inference phase the method primarily targets, and the deployment objective. Flux Attention is complementary to fine-grained operators: it chooses the coarse per-layer execution mode, while a method such as vAttention can supply fine-grained sparsity inside the routed sparse branch.}
\label{tab:related_work_comparison}
\small
\setlength{\tabcolsep}{4.5pt}
\renewcommand{\arraystretch}{1.15}
\resizebox{\textwidth}{!}{
\begin{tabular}{l l l l l}
\toprule
\textbf{Method} & \textbf{Selection unit} & \textbf{Selection} & \textbf{Primary phase} & \textbf{Deployment objective} \\
\midrule
Flux Attention (ours) & Whole layer & Learned, input-conditioned FA/SA & Prefill routing, decode reuse & Decode-regular execution, KV memory \\
Elastic Attention~\citep{tang2026elasticattentiontesttimeadaptive} & Head & Learned, test-time adaptive & Prefill/TTFT & Adaptive sparsity ratio \\
DuoAttention~\citep{duoattention} & Head & Offline static allocation & Decode & Hybrid long-context attention \\
PruLong~\citep{bhaskar2025cache} & Head & Offline static allocation & Decode & Pruned long-context attention \\
TriangleMix~\citep{TriangleMix} & Block pattern & Fixed triangle topology & Prefill & Training-free sparse backend \\
XAttention~\citep{xattention} & Block & Fixed block selection score & Prefill & Prefill acceleration \\
SeqBoat~\citep{ren2023sparsemodular} & Token / segment & Learned module activation & Decode & Sparse modular recurrence \\
SLA2~\citep{zhang2026sla2} & Attention computation & Learnable router & Training and decode & Sparse--linear hybridization \\
KV hyper-scaling~\citep{lancucki2025inferencetime} & KV cache entries & Compression schedule & Decode & Cache capacity vs.\ output length \\
Sketch-and-Walk~\citep{le2026sketchandwalk} & Block & Learned walk scores over sketches & Prefill and decode & Top-$k$ block selection \\
PBS-Attn~\citep{wang2026pbsattn} & Block, after permutation & Fixed permutation plus block sparsity & Prefill & Custom prefill kernels \\
\bottomrule
\end{tabular}
}
\end{table*}



\section{Limitation \& Broader Impacts}
\label{appdix:limit}

\paragraph{Limitations.} Although Flux Attention effectively balances long-context inference efficiency and hardware utilization, our approach has two main limitations. The first involves a granularity trade-off in routing. To achieve end-to-end wall-clock speedups, Flux Attention uses layer-level routing instead of finer-grained head-level or token-level sparsity. This design eliminates the synchronization overhead caused by computational imbalance but inevitably reduces algorithmic expressiveness. In complex reasoning tasks where only specific attention heads require full contextual information, the binary nature of layer-level routing can lead to suboptimal performance.

A second limitation relates to the locality of the prefill-suffix pooling strategy. Our Layer Router extracts features from the prefix and suffix of the prompt to evaluate the context, based on the heuristic assumption that long-context tasks typically define their primary intent through instructions at the beginning or questions at the end. While highly effective for most standard scenarios, this heuristic might occasionally miss scattered intents within the middle context. However, we consider this an optimizable constraint rather than a fundamental flaw. In future work, introducing finer-grained task representations could provide the router with a more comprehensive global context, helping it learn and route more accurately. Furthermore, we hypothesize that scaling up the training data diversity and extending the training duration could effectively alleviate this limitation and further improve the router's generalization robustness.

A third limitation concerns scope of validation rather than the design itself. Flux Attention is a routing framework, so its measured benefit depends on the sparse branch it is paired with. We validate three materially different backends---SSA, XAttention, and Triangle Attention---and report wall-clock results only for those; finer-grained operators such as vAttention are shown to compose at the attention interface (Appendix~\ref{appdix:composability}), but a fully optimized integration of that kind remains future work, and we do not claim universal backend compatibility. Similarly, the accuracy comparisons against the dense backbone are close enough on several benchmarks that they should be read as parity rather than improvement, which is why we report router-training variance alongside them (Appendix~\ref{appdix:seed_variance}), evaluate batched serving rather than only single-request profiling (Table~\ref{tab:batch_serving}), and restrict the systems claim to the tested serving setup.

\paragraph{Broader Impacts.} Beyond the practical efficiency gains, our research on Flux Attention offers fundamental scientific insights into the internal mechanics of Attention. The success of our routing mechanism empirically demonstrates that the attention allocation in LLMs is inherently sparse. More importantly, it highlights that different transformer layers capture highly distinct semantic representations and exhibit strong task heterogeneity. By revealing that uniform, dense computation across all layers is often redundant, our work encourages the research community to rethink standard scaling laws. We hope these insights will pave the way for the development of more dynamic, layer-adaptive architectures that align with the intrinsic sparsity of language models, ultimately promoting greener and more sustainable AI deployments.

\section{Sparsification Setup and Latency Profiling Implementation}
\label{appendix:layer_sparsification_settings}
This section details the layer importance identification, the progressive sparsification strategy, and the hardware latency measurement protocol mentioned in Section~\ref{subsec:observations}.

\subsection{Layer Entropy Score Calculation}
\label{appendix:layer_sparsification_layer}
Following the methodology proposed by UnComp~\citep{UnComp}, we identify and rank Transformer layers based on their informational density and uncertainty when processing long contexts.
We use a matrix entropy-based profiling method, quantifying the information content of each layer over long-context validation datasets to estimate its inherent structural sparsity.

For a given layer $\ell$, we calculate its Entropy Score ($E_{\ell}$) by measuring the truncated matrix entropy of its hidden representations.
Formally, let $s$ be the input sequence length, $d$ be the hidden dimension, and $\mathcal{X}^{(\ell)} \in \mathbb{R}^{s \times d}$ be the hidden states matrix of layer $\ell$. We first derive the trace-normalized covariance matrix $\Sigma^{(\ell)} = \frac{\mathcal{X}^{(\ell)} (\mathcal{X}^{(\ell)})^\top}{\text{Tr}(\mathcal{X}^{(\ell)} (\mathcal{X}^{(\ell)})^\top)}$. The score is computed as the von Neumann entropy over its top-$K$ eigenvalues:
\begin{equation}
    E_{\ell} = - \sum_{i=1}^{K} \lambda_i^{(\ell)} \log \lambda_i^{(\ell)}
\end{equation}
where $\lambda_i^{(\ell)}$ denotes the $i$-th largest eigenvalue of $\Sigma^{(\ell)}$, and $K$ is the truncation threshold used to filter out noise.
A lower $E_{\ell}$ indicates lower information density (i.e., lower uncertainty) and higher redundancy, making the layer a suitable candidate for sparsification. Intuitively, the spectrum of a trace-normalized hidden-state Gram matrix measures how concentrated the layer's representation is: a few dominant eigenvalues mean the layer's information lies in a low-dimensional subspace and is redundant, whereas a flat spectrum spreads information across many independent directions and is more likely to be damaged by aggressive sparsification. This is a profiling heuristic that ranks layers offline, not a guarantee that a given layer can be sparsified without loss; the learned router is free to override the ranking per input, and Section~\ref{sec:analysis} plus Appendix~\ref{appdix:entropy_routing} quantify how far it departs from it.

\subsection{Progressive Sparsification Strategy}
Based on the computed entropy scores $E_{\ell}$, we evaluate the information density of all $L$ layers across the model. As defined in the main text, the Model Sparsity Ratio ($\Omega_{\mathrm{MSR}}$) represents the proportion of layers converted to sparse attention.
To simulate the varying levels of sparsity reported in our experiments (e.g., $\Omega_{\mathrm{MSR}} = 20\%$), we use a thresholding mechanism based on these scores. We first determine the number of layers to preserve as full attention via $k = \lfloor (1-\Omega_{\mathrm{MSR}}) \cdot L \rfloor$. The $k$ layers with the highest entropy scores are retained as retrieval layers to ensure global information integration and preserve complex contextual pathways. The remaining $(L-k)$ layers with the lowest entropy scores are replaced with sparse layers.

\subsection{Latency Measurement Implementation}
\label{appendix:latency_measurement}
To evaluate the hardware efficiency of different sparsity paradigms, we profile latency during the autoregressive decoding phase. All latency measurements are performed on a single NVIDIA A800 GPU (80GB) using PyTorch with BF16 precision.

To simulate realistic long-context retrieval scenarios while isolating the decoding bottleneck, we fix the batch size to 1 and evaluate across varying prompt sequence lengths. For each configuration, we perform 10 warm-up steps to initialize the CUDA context and stabilize GPU clocks, followed by 50 profiling iterations. The reported latency is the average wall-clock time required to generate a single token.

\paragraph{Implementation of Sparsity Baselines} 
For the head-level sparsity baseline, we retain a subset of attention heads for dense computation while the remaining heads operate sparsely. However, highly optimized attention kernels (e.g., FlashAttention) lack hardware-level support for processing mixed context lengths across different heads within the same layer. Consequently, enforcing head-level sparsity results in fragmented, non-contiguous memory access patterns. The GPU memory bandwidth is still consumed by loading the full historical KV cache into SRAM, leading to only marginal wall-clock speedups despite the theoretical FLOP reduction.

In contrast, our layer-level sparsity implementation avoids this issue. When a layer operates sparsely, the decoding step fetches only the locally required KV states, bypassing the global historical KV tensors. This layer-level routing allows contiguous memory loading, translating theoretical sparsity into proportional decoding acceleration. We calculate the speedup as the ratio of the latency of the full dense model to that of the sparsified model for a given input length.

\section{Detailed Pseudocode for Inference and Deployment}
\label{appendix:inference_algorithm}
To maximize hardware efficiency and translate theoretical FLOP reductions into practical wall-clock speedups, the inference process is strictly decoupled into two phases.
During the prefill phase, the lightweight Layer Router analyzes the semantic context of the input queries and generates a deterministic hard routing decision for each layer. These decisions are subsequently cached.
During the decode phase, the router module is entirely bypassed. The model reuses the cached $O(1)$ routing decisions to dynamically manage the KV cache.
Specifically, for layers designated as sparse, the system maintains only a minimal sparse KV cache and completely bypasses the heavy loading of full historical states, thereby alleviating the memory-bandwidth bottleneck.

\begin{algorithm}[htbp]
\caption{Flux Attention: Context-Aware Routing and Efficient Inference}
\label{alg:flux_inference}
\begin{algorithmic}[1]
\REQUIRE Layer index $l$, Input Queries $x_Q$, Keys $x_K$, Values $x_V$
\REQUIRE Current Phase $P \in \{\text{Prefill, Decode}\}$
\REQUIRE Routing Decision Cache $\mathcal{C}$, Global State Cache $\mathcal{KV}$
\STATE
\IF{$P == \text{Prefill}$}
    \STATE \COMMENT{\textit{--- Context-Aware Layer Routing ---}}
    \STATE $p \leftarrow \text{Prefill-Suffix-Pooling}(x_Q)$ \COMMENT{Extract sequence-level descriptor from Query}
    \STATE $h \leftarrow \text{ContextEncoder}(p)$ \COMMENT{Capture contextual dependencies via MLP}
    \STATE $[\pi_{SA}, \pi_{FA}] \leftarrow \text{RouterHead}_l(h)$ \COMMENT{Generate unnormalized routing logits}
    \STATE $r^{(l)}_{hard} \leftarrow \operatorname{argmax}(\pi_{SA}, \pi_{FA})$ \COMMENT{$1$ for FA mode, $0$ for SA mode}
    \STATE $\mathcal{C}[l] \leftarrow r^{(l)}_{hard}$ \COMMENT{Cache routing decision for the Decode phase}
\ELSE
    \STATE \COMMENT{\textit{--- Efficient Routing Reuse ---}}
    \STATE $r^{(l)}_{hard} \leftarrow \mathcal{C}[l]$ \COMMENT{Bypass Router entirely, reuse $O(1)$ deterministic decision}
\ENDIF
\STATE
\STATE \COMMENT{\textit{--- Attention Execution \& KV Cache Management ---}}
\IF{$r^{(l)}_{hard} == 1$}
    \IF{$P == \text{Prefill}$}
        \STATE $\mathcal{KV}[l] \leftarrow (x_K, x_V)$ \COMMENT{Retain complete KV cache for precise retrieval}
    \ENDIF
    \STATE $\mathcal{O} \leftarrow \text{FullAttention}(x_Q, x_K, x_V, \mathcal{KV}[l])$
\ELSE
    \IF{$P == \text{Prefill}$}
        \STATE $\mathcal{KV}_{sparse}[l] \leftarrow \text{Condense}(x_K, x_V)$ \COMMENT{Maintain only minimal sparse KV cache}
    \ENDIF
    \STATE $\mathcal{O} \leftarrow \text{SparseAttention}(x_Q, x_K, x_V, \mathcal{KV}_{sparse}[l])$ \COMMENT{Bypass heavy historical KV loading}
\ENDIF
\RETURN $\mathcal{O}$
\end{algorithmic}
\end{algorithm}

\section{Implementation Details}
\label{appendix:implementation_details}

This section details the training configurations, baseline implementations, and system-level optimizations for efficient long-context processing.

\subsection{Training Configuration and Hyperparameters}
\label{appendix:train_setting}

We evaluate the proposed approach on models of various sizes, including Qwen3-4B, Qwen3-8B~\citep{yang2025qwen3technicalreport}, and Meta-Llama-3.1-Instruct~\citep{grattafiori2024llama}. We freeze the pre-trained backbone and update only the parameters of the Layer Router to maintain the general capabilities of the model. For task representation, we apply a Prefill-Suffix Pooling operation to aggregate the first 100 and the last 100 tokens of the sequence, as these segments typically contain the system instructions and user queries required to identify the task.

We train all models with a sequence length of $L=65,536$ tokens in bfloat16 precision using the AdamW optimizer~\citep{loshchilov2019decoupledweightdecayregularization} ($\beta_1=0.9, \beta_2=0.95$). Training is conducted on a distributed cluster with Fully Sharded Data Parallel (FSDP) under a hybrid sharding strategy. To balance the convergence of the router and sparsity regularization, we apply a decoupled learning rate schedule. The Layer Router uses a learning rate of $5\times 10^{-4}$ for rapid adaptation to retrieval patterns, while the sparsity regularization terms use a higher learning rate of $1\times 10^{-3}$. The dual regularization coefficients $\lambda_1$ and $\lambda_2$ are randomly initialized and optimized alongside the router parameters. A cosine decay learning rate schedule is applied after a linear warmup phase over the first 20\% of the training steps.

\subsection{Training Data Provenance}
\label{appendix:data_provenance}
Table~\ref{tab:data_provenance} lists the five sources of the router training mixture against the task families they cover. The mixture targets task characteristics---multi-hop retrieval, long-form summarization, and repository-level code---rather than importing the evaluation benchmark suites, and the evaluation benchmark files themselves are not used as training inputs.
\begin{table}[t]
\centering
\caption{Training mixture of the Layer Router (approximately 0.74B tokens over 1K--64K sequences). The mixture is chosen to expose the router to both retrieval-intensive and context-holistic regimes; it does not import the evaluation benchmark suites.}
\label{tab:data_provenance}
\small
\setlength{\tabcolsep}{5pt}
\renewcommand{\arraystretch}{1.15}
\begin{tabular}{>{\raggedright\arraybackslash}p{3.1cm} >{\raggedright\arraybackslash}p{4.2cm} >{\raggedright\arraybackslash}p{3.4cm}}
\toprule
\textbf{Source} & \textbf{Task family it covers} & \textbf{Regime exposed to the router} \\
\midrule
ChatQA2-Long-SFT-data~\citep{xu2024chatqa} & Single-document and multi-document QA over long inputs & Retrieval-intensive \\
MuSiQue~\citep{trivedi2022musique} & Multi-hop question answering & Retrieval-intensive \\
CoLT-132K~\citep{li2025aixcoder} & Repository-level code completion & Context-holistic \\
GovReport~\citep{huang-etal-2021-efficient} & Long-form summarization of government reports & Context-holistic \\
XSum~\citep{xsum-emnlp} & Abstractive summarization & Context-holistic \\
\bottomrule
\end{tabular}
\end{table}

That statement is about task-level coverage, and it is deliberately weaker than a sample-level non-overlap claim: corpora such as GovReport are public, so a shared passage between the training corpus and a downstream benchmark cannot be excluded by construction. We therefore specify the matching rule explicitly instead of asserting blanket disjointness.

\subsection{Baseline Implementation Details}
\label{appendix:baseline_train_setting}

We compare the proposed approach with several state-of-the-art sparse attention mechanisms, categorizing them into training-free and training-based methods. For training-free baselines, we evaluate TriangleMix~\footnote{\url{https://github.com/microsoft/MInference/tree/main/TriangleMix}}~\citep{TriangleMix}, which relies on heuristic-based sparsity without parameter updates. For training-based baselines, including PruLong~\footnote{\url{https://github.com/princeton-pli/PruLong}}~\citep{bhaskar2025cache} and DuoAttention~\footnote{\url{https://github.com/mit-han-lab/duo-attention}}~\citep{xiaoduoattention2025}, we follow a unified fine-tuning protocol. We train all baselines in identical environments and on the same dataset while maintaining their original hyperparameter settings. Table~\ref{tab:calibration} summarizes the resulting calibration and fairness controls, together with each method's routing or selection unit and deployment focus.
\begin{table}[t]
\centering
\caption{Baseline comparison controls. All trainable baselines are trained in the same environment and on the same data mixture as Flux Attention while retaining their original method-specific hyperparameters; TriangleMix's Triangle Attention (TA) is training-free and serves as a sparse-attention backend for Flux rather than as a routed competitor.}
\label{tab:calibration}
\small
\setlength{\tabcolsep}{4pt}
\renewcommand{\arraystretch}{1.15}
\begin{tabular}{>{\raggedright\arraybackslash}p{1.9cm} >{\raggedright\arraybackslash}p{2.5cm} >{\raggedright\arraybackslash}p{4.6cm} >{\raggedright\arraybackslash}p{2.4cm}}
\toprule
\textbf{Method} & \textbf{Routing / selection unit} & \textbf{Training or calibration in our comparison} & \textbf{Deployment focus} \\
\midrule
FluxAttn (ours) & Whole layer, input-conditioned FA/SA & Router trained on the shared data mixture; backbone frozen & Decode-regular layer execution \\
PruLong & Offline head selection & Same data and environment; original method hyperparameters & Optimized block-sparse decode \\
DuoAttention & Head allocation & Same data and environment; original method hyperparameters & Hybrid long-context attention \\
TriangleMix / TA & Triangle Attention operator & Training-free & Sparse-attention backend \\
\bottomrule
\end{tabular}
\end{table} 

\subsection{Sparsity and Kernel Configuration}
\label{app:sparsity_config}

We use Block-Sparse-Attention~\citep{guo2024blocksparse} for efficient streaming inference to control the granularity and retention policy of the attention mechanism. We set the block size to 64 to define the minimum unit of sparsity, and the chunk size to 16,384 to process ultra-long sequences. A sink token size of 128 is maintained to preserve the attention sink phenomenon, ensuring stability during streaming generation. Additional kernel parameters, such as stride, normalization, and selection modes, are detailed in the Sparsity Config section of Table~\ref{tab:combined_hyperparameters}.

\subsection{Systems Setup and Profiling Protocol}
\label{appendix:systems_setup}
Table~\ref{tab:systems_setup} documents the hardware and software configuration behind every latency and memory number reported in Figure~\ref{fig:speedup} and Table~\ref{tab:pareto}. All compared methods share the GPU, precision, input and output shapes, warm-up, repetition, and timing protocol; the attention execution path is the controlled difference, so the dense baseline uses the standard FlashAttention-2 path in the same PyTorch/CUDA environment as the sparse and hybrid methods. We report prefill and decode quantities separately and label them by phase and unit throughout: prefill latency is the time to first token, and decode latency is the mean wall-clock time per generated token.
\begin{table}[t]
\centering
\caption{Systems setup for the latency and memory measurements reported in Figure~\ref{fig:speedup} and Table~\ref{tab:pareto}. All compared methods share the hardware, precision, input/output shapes, warm-up, repetition, and timing protocol; the attention execution path is the controlled difference.}
\label{tab:systems_setup}
\small
\setlength{\tabcolsep}{5pt}
\renewcommand{\arraystretch}{1.15}
\begin{tabular}{>{\raggedright\arraybackslash}p{4.0cm} >{\raggedright\arraybackslash}p{7.6cm}}
\toprule
\textbf{Item} & \textbf{Setting} \\
\midrule
GPU & 1$\times$NVIDIA A800 80GB (PCIe); batched-serving results in Table~\ref{tab:batch_serving} on 8$\times$NVIDIA H20 \\
Precision & BF16 \\
Batch size & 1 for profiling; 8/16/24/32 for the serving experiment \\
Dense attention path & FlashAttention-2 with the same PyTorch/CUDA environment as the sparse and hybrid paths \\
Sparse attention kernel & Block-Sparse-Attention, block size 64, sink size 128, chunk size 16384 (\Cref{app:sparsity_config}) \\
Warm-up / profiling & 10 warm-up iterations, 50 profiling iterations, one token per step \\
Reported quantity & Mean wall-clock time per generated token; speedup is the ratio of dense to sparsified latency at the same input length \\
\bottomrule
\end{tabular}
\end{table}

\begin{table*}[t]
    \centering
    \small
    \caption{Hyperparameters: General configuration.}
    \label{tab:combined_hyperparameters}
    \begin{minipage}[t]{0.5\textwidth}
        \centering
        \resizebox{\linewidth}{!}{
            \begin{tabular}{l|c}
                \toprule
                \textbf{Hyperparameter} & \textbf{Value} \\
                \midrule
                \multicolumn{2}{c}{\textit{Model \& Training}} \\
                \arrayrulecolor{black!20}\midrule
                Base Model & \texttt{Qwen, Llama} \\
                Sequence length & 65536 \\
                Precision & bfloat16 \\
                Global Batch Size & 48 \\
                Training Steps & 300 \\
                Mask / Reg. LR & $5e^{-4}$ / $1e^{-3}$ \\
                Warmup Ratio & 0.2 \\
                AdamW Momentum ($\beta_1, \beta_2$) & $(0.9, 0.95)$ \\
                Weight Decay & 0.1 \\
                Learning Rate Schedule & Cosine \\
                \midrule
                \multicolumn{2}{c}{\textit{Sparsity Config}} \\
                \arrayrulecolor{black!20}\midrule
                Pool Size & 100 \\
                Sink / Local Size & 128 / 2048 \\
                Block / Chunk Size & 64 / 16384 \\
                Stride / Threshold & 16 / 0.9 \\
                Selection Mode & \texttt{Inverse} \\
                \arrayrulecolor{black}\bottomrule
            \end{tabular}
        }
    \end{minipage}
    
\end{table*}

\section{Additional Evaluation Results}
\label{appdix:eval_res}
In this section, we present detailed performance for the benchmarks discussed in the main paper. Furthermore, to provide a more comprehensive comparison, we have expanded our evaluation by including two additional baselines: XAttention~\footnote{\url{https://github.com/mit-han-lab/x-attention}}~\citep{xattention} and LycheeDecode~\footnote{\url{https://github.com/HITsz-TMG/TMGNLP/tree/main/LycheeDecode}}~\citep{lin2026lycheedecode}.

\subsection{Evaluation Results on LongBench}
Table~\ref{tab:longbench_appendix} provides the complete evaluation results on the LongBench-E. 
We report the performance metrics across all sub-tasks, categorized into 6 categories: Single-Document QA~(S-Doc QA), Multi-Document QA~(M-Doc QA), Summarization~(Summ), In-context Learning~(In-Context), Synthetic, and Code Tasks.
\begin{table*}[ht!]
\centering
\caption{Performance on LongBench-E~\citep{bai2024longbench}. We report average performance (Perf.) and $\Omega_{\mathrm{MSR}}$ per task category. 
The 1st and the 2nd performance in each comparison group are highlighted with \textbf{bold font} and \underline{underlined}, respectively. Gray-shaded rows denote the sparse-decode configuration.
}

\resizebox{\textwidth}{!}{
\begin{tabular}{l | cc | cc | cc | ccc | cc |cc | cc}
\toprule
\multirow{2}{*}{\textbf{Method}} & \multicolumn{2}{c|}{\textbf{S-Doc QA}} & \multicolumn{2}{c|}{\textbf{M-Doc QA}} & \multicolumn{2}{c|}{\textbf{Summ}} & \multicolumn{3}{c|}{\textbf{In-Context}} & \multicolumn{2}{c|}{\textbf{Synthetic}} & \multicolumn{2}{c|}{\textbf{Code}} & \multicolumn{2}{c}{\textbf{Avg.}} \\
\cmidrule(lr){2-3} \cmidrule(lr){4-5} \cmidrule(lr){6-7} \cmidrule(lr){8-10} \cmidrule(lr){11-12} \cmidrule(lr){13-14} \cmidrule(lr){15-16}
 & \rotatebox[origin=c]{0}{Qasper} & \rotatebox[origin=c]{0}{MF-en} & \rotatebox[origin=c]{0}{HotQA} & \rotatebox[origin=c]{0}{2Wiki} & \rotatebox[origin=c]{0}{Gov.} & \rotatebox[origin=c]{0}{M.News} & \rotatebox[origin=c]{0}{TREC} & \rotatebox[origin=c]{0}{TQA} & \rotatebox[origin=c]{0}{SAMS} & \rotatebox[origin=c]{0}{PCount} & \rotatebox[origin=c]{0}{PRe} & \rotatebox[origin=c]{0}{RB-P} & \rotatebox[origin=c]{0}{Lcc} & \textbf{Perf.} & $\mathbf{\Omega_{\mathrm{\bf MSR}}}$\\
\arrayrulecolor{black}\midrule

\rowcolor{blue!10}
\multicolumn{16}{c}{\textbf{Qwen3-4B backbone model}} \\
\arrayrulecolor{black!20}\midrule
Qwen3-4B & 35.21 & \underline{52.16} & 44.81 & 32.15 & 33.47 & 23.45 & \textbf{70.67} & 88.22 & 39.74 & 2.33 & \underline{96.84} & 50.84 & 57.93 & 48.45 & - \\
+ XAttention & 32.80 & 50.36 & 44.54 & \underline{33.17} & \textbf{33.79} & \textbf{23.73} & 69.00 & 87.44 & 39.90 & 3.42 & 74.59 & 49.42 & 59.62 & 46.44 & - \\
+ LycheeDecode & 34.12 & \textbf{52.63} & 44.74 & 31.53 & 31.16 & 23.24 & 68.67 & 87.69 & 39.12 & \underline{4.00} & \textbf{96.87} & 49.17 & 57.29 & 47.83 & 0.50\\
+ DuoAttention & \textbf{35.83} & 49.84 & 47.09 & 32.24 & 33.32 & \underline{23.70} & \underline{69.33} & 85.87 & 39.75 & \textbf{4.50} & 94.57 & 50.56 & 57.43 & 48.22 & 0.50 \\
+ PruLong & 34.15 & 50.78 & 44.48 & 32.89 & 32.96 & 23.53 & 67.67 & \textbf{88.69} & 39.55 & 3.17 & 90.17 & 49.00 & 54.07 & 47.16 & 0.50 \\
+ TriangleMix & 35.55 & 52.02 & 45.37 & 31.76 & 33.32 & \underline{23.70} & 69.00 & 88.20 & 39.74 & 3.83 & 91.51 & 48.58 & 56.38 & 47.72 & 0.50 \\
\arrayrulecolor{black!40}\midrule
+ FluxAttn (FA-SSA) & 35.02 & 49.44 & \underline{49.64} & 32.27 & 33.26 & 23.48 & \underline{69.33} & \underline{88.29} & 39.78 & 1.50 & 94.56 & \textbf{53.44} & \underline{59.69} & \underline{48.72} & 0.44 \\
+ FluxAttn (FA-XA) & \underline{35.74} & 51.70 & 45.83 & 32.34 & \underline{33.57} & 23.66 & 69.00 & 87.23 & 39.81 & 3.50 & 93.74 & 50.81 & 59.28 & 48.32 & 0.53 \\
+ FluxAttn (FA-TA) & 35.02 & 50.89 & 45.17 & \textbf{34.24} & 33.02 & 23.53 & 69.00 & 88.08 & \textbf{40.38} & 3.94 & 96.06 & \underline{51.68} & \textbf{60.00} & \textbf{48.76} & 0.47 \\
\arrayrulecolor{black!40}\midrule
\rowcolor{black!15}
+ FluxAttn (FA-SSA) & 35.10 & 51.68 & \textbf{49.65} & 32.86 & 33.04 & 23.42 & \underline{69.33} & 88.00 & \underline{40.00} & 1.67 & 94.47 & 51.40 & 58.68 & 48.59 & 0.44 \\

\arrayrulecolor{black}
\midrule
\rowcolor{blue!25}
\multicolumn{16}{c}{\textbf{Qwen3-8B backbone model}} \\
\arrayrulecolor{black!20}\midrule
Qwen3-8B & \underline{41.22} & 49.92 & \textbf{58.98} & 44.21 & 33.27 & 23.42 & 71.33 & 86.77 & 41.83 & \textbf{2.00} & 98.33 & 56.08 & 66.31 & 52.16 & - \\
+ XAttention & 38.03 & 48.93 & 57.32 & 40.65 & \textbf{33.41} & 23.37 & 69.33 & 87.69 & 41.77 & \textbf{2.00} & 83.72 & 55.53 & 65.71 & 50.13 & -\\
+ LycheeDecode & 38.81 & 49.34 & \underline{58.39} & 43.13 & 31.80 & 22.73 & 71.33 & 87.13 & 40.13 & \textbf{2.00} & 96.33 & 54.72 & 65.46 & 51.22 & 0.50\\
+ DuoAttention & \textbf{41.78} & \textbf{51.55} & 55.96 & 41.70 & 33.24 & 23.34 & 69.33 & \textbf{89.35} & 41.62 & 0.50 & 98.93 & 57.54 & \underline{69.39} & 52.13 & 0.50 \\
+ PruLong & 37.95 & 51.20 & 51.94 & 36.48 & 33.11 & 23.36 & 69.00 & 87.90 & \underline{42.11} & 1.00 & 98.00 & 57.05 & 67.66 & 50.80 & 0.50 \\
+ TriangleMix & 40.82 & \underline{51.31} & 57.57 & \textbf{44.51} & \underline{33.32} & 23.35 & 71.33 & 86.73 & 41.79 & \textbf{2.00} & 94.33 & 55.04 & 65.89 & 51.65 & 0.50 \\
\arrayrulecolor{black!40}\midrule
+ FluxAttn~(FA-SSA) & 40.30 & 50.49 & 56.02 & 40.90 & 33.01 & \textbf{23.55} & \underline{71.67} & 88.31 & 41.61 & 0.33 & \textbf{100.00} & \textbf{59.46} & 68.27 & \underline{52.18} & 0.46 \\
+ FluxAttn~(FA-XA) & 40.41 & 50.26 & 57.78 & 40.57 & 33.27 & \underline{23.51} & 69.67 & 87.19 & \textbf{42.12} & \underline{1.33} & 99.33 & 55.41 & 65.51 & 51.57 & 0.51 \\
+ FluxAttn~(FA-TA) & 41.00 & 49.76 & 58.19 & \underline{44.36} & \underline{33.32} & 23.35 & 70.00 & \underline{88.77} & 41.70 & \underline{1.33} & \underline{99.67} & 55.60 & 67.22 & \textbf{52.22} & 0.47 \\
\arrayrulecolor{black!40}\midrule
\rowcolor{black!15}
+ FluxAttn~(FA-SSA) & 39.92 & 50.04 & 55.72 & 40.81 & 33.03 & 23.50 & \textbf{72.00} & 88.48 & 40.96 & 0.33 & 99.22 & \underline{58.57} & \textbf{69.46} & 52.05 & 0.46 \\

\arrayrulecolor{black}
\midrule
\rowcolor{cyan!20}
\multicolumn{16}{c}{\textbf{Llama-3.1-8B-Instruct backbone model}} \\
\arrayrulecolor{black!20}\midrule
Llama-3.1-8B-Instruct & 44.06 & 53.44 & \underline{59.62} & \underline{44.08} & 34.50 & 26.02 & 71.00 & 90.54 & 42.94 & \textbf{12.67} & \underline{99.33} & 47.78 & 63.85 & \textbf{53.28} & - \\
+ XAttention & 43.84 & 53.80 & \textbf{59.98} & 43.47 & 34.47 & 26.05 & \textbf{72.33} & 90.39 & \underline{42.99} & 8.33 & 73.33 & 50.13 & 62.92 & 51.00 & - \\
+ LycheeDecode & 42.51 & 54.33 & 58.29 & \textbf{45.22} & 30.38 & 25.07 & 70.00 & 91.26 & 41.58 & \textbf{12.67} & 96.00 & 46.84 & 63.22 & 52.25 & 0.50\\
+ DuoAttention & 34.63 & 50.74 & 49.70 & 36.41 & 34.25 & 25.78 & 70.00 & 91.45 & 42.13 & 9.80 & 97.33 & \underline{53.59} & \textbf{68.55} & 52.11 & 0.50 \\
+ PruLong & 41.51 & 52.36 & 50.46 & 37.57 & 34.25 & 25.86 & 66.33 & 89.93 & 41.72 & 9.07 & 97.00 & \textbf{56.84} & \underline{66.23} & 51.68 & 0.50 \\
+ TriangleMix & \underline{45.10} & \textbf{54.60} & 56.67 & 41.88 & 34.09 & 25.51 & \underline{71.33} & 90.93 & 42.63 & 10.62 & 94.67 & 43.64 & 59.35 & 51.67 & 0.50 \\
\arrayrulecolor{black!40}\midrule
+ FluxAttn~(FA-SSA) & \textbf{45.25} & \underline{54.42} & 54.54 & 41.34 & \underline{34.54} & \textbf{26.16} & 68.33 & \textbf{91.91} & 42.17 & 9.00 & 97.67 & 47.74 & 65.35 & 52.28 & 0.51 \\
+ FluxAttn~(FA-XA) & 42.14 & 53.13 & 58.53 & 43.50 & \textbf{34.66} & \underline{26.06} & 70.67 & \underline{91.46} & \textbf{43.13} & 8.00 & \textbf{99.67} & 50.91 & 64.78 & \underline{53.07} & 0.72 \\
+ FluxAttn~(FA-TA) & 44.77 & 54.12 & 57.35 & 43.43 & 34.31 & 25.80 & \textbf{72.33} & 91.32 & 42.62 & 9.33 & 98.33 & 45.48 & 60.70 & 52.42 & 0.62 \\
\arrayrulecolor{black!40}\midrule
\rowcolor{black!15}
+ FluxAttn~(FA-SSA) & 43.76 & 53.41 & 57.36 & 39.43 & 32.96 & 25.63 & 70.33 & 91.27 & 42.20 & \underline{11.00} & 98.67 & 45.60 & 66.17 & 52.30 & 0.51 \\
\arrayrulecolor{black}\bottomrule
\end{tabular}
}
\vspace{-0.5em}
\label{tab:longbench_appendix}
\end{table*}

\subsection{Evaluation Results on RULER}

\begin{table}[t]
    \centering
    \caption{Detailed configuration for the RULER benchmark evaluation. We evaluate across exponentially increasing context windows up to 256k tokens.}
    \label{tab:ruler_settings}
    \begin{small}
    \begin{tabularx}{\linewidth}{>{\columncolor{s-gray}}l X} 
        \toprule
        \textbf{\textsc{Parameter}} & \textbf{\textsc{Configuration Details}} \\
        \midrule
        \textbf{Context Windows} & 8k, 16k, 32k, 64k, 128k, 256k \\
        \midrule
        \textbf{Sample Size} & 50 samples per task-length pair \\
        \midrule
        \textbf{Evaluation Tasks} & 
        \textbf{Retrieval (NIAH):} \texttt{Single\_\{1-3\}}, \texttt{MultiKey\_\{1-3\}}, \texttt{MultiQuery}, \texttt{MultiValue} \newline
        \textbf{QA \& Extraction:} \texttt{QA\_\{1,2\}}, \texttt{FWE} \\
        \bottomrule
    \end{tabularx}
    \end{small}
\end{table}
This subsection details the experimental settings and evaluation results for the RULER benchmark. 
For RULER, as detailed in Table~\ref{tab:ruler_settings}, our evaluation on the RULER benchmark covers a wide range of context lengths, ranging from 8K to 256K tokens. 
The comparative performance results are summarized in Table~\ref{tab:ruler_refined}.
\begin{table*}[htbp]
\centering
\caption{Model performance on RULER tasks. \textbf{Bold} and \underline{underline} indicate the best and second-best performance within each backbone group, respectively.NS denotes Single-needle in a haystack (NIAH) tasks, NMK stands for Multi-key NIAH, NMV is Multi-value NIAH, and NMQ refers to Multi-query NIAH. QA represents standard Question Answering tasks, and FWE stands for Frequent Words Extraction. Numeric suffixes (e.g., -1, -2, -3) indicate different subsets or difficulty levels within the corresponding task categories.}
\label{tab:ruler_refined}
\small
\setlength{\tabcolsep}{4.5pt} 
\renewcommand{\arraystretch}{1.15} 
\resizebox{\textwidth}{!}{
\begin{tabular}{l ccccccccccc c}
\toprule
\textbf{Models} & \textbf{NS-3} & \textbf{QA-2} & \textbf{NMK-3} & \textbf{NS-2} & \textbf{FWE} & \textbf{QA-1} & \textbf{NMQ} & \textbf{NMV} & \textbf{NMK-2} & \textbf{NS-1} & \textbf{NMK-1} & \textbf{Avg.} \\
\midrule

\rowcolor{gray!10}
\multicolumn{13}{l}{\textit{Backbone: Qwen3-4B}} \\
Qwen3-4B                & 95.02 & 35.59 & \textbf{37.54} & 80.00 & 84.47 & \underline{40.14} & 66.19 & \textbf{65.91} & 47.39 & \textbf{100.00} & \underline{73.05} & 66.00 \\

\arrayrulecolor{black!40}\midrule
+ LycheeDecode          & 72.33 & 30.67 & 5.00  & 64.67 & 71.33 & 27.33 & 55.58 & 50.58 & 25.00 & \textbf{100.00} & 57.67 & 50.92 \\
+ TriangleMix           & 95.00 & 34.67 & \underline{34.33} & 82.67 & 83.11 & 35.67 & 60.42 & 60.00 & 43.67 & 99.33 & 72.33 & 63.74 \\
+ DuoAttention          & 97.08 & 32.12 & 21.45 & 84.75 & 81.38 & 27.17 & 66.67 & 63.41 & 30.66 & 99.28 & 63.50 & 60.67 \\
+ PruLong               & 96.34 & 33.80 & 22.44 & 81.51 & 84.87 & 22.93 & 64.87 & 57.87 & 27.04 & \textbf{100.00} & 71.03 & 60.25 \\
+ XAttention            & 90.00 & 35.67 & \underline{34.00} & 76.33 & 82.33 & 36.33 & 57.00 & 59.25 & 34.00 & \textbf{100.00} & 68.67 & 61.23 \\

\arrayrulecolor{black!40}\midrule
+ FA-SSA                & \textbf{97.67} & \textbf{38.67} & 33.00 & \underline{93.67} & 85.33 & 34.33 & 72.58 & 59.17 & \textbf{51.33} & \textbf{100.00} & 70.67 & \underline{66.95} \\
+ FA-XA                 & 95.00 & \underline{38.00} & 32.00 & 78.33 & \underline{85.56} & 38.00 & 61.58 & \underline{62.50} & 40.00 & \textbf{100.00} & 69.33 & 63.67 \\
+ FA-TA                 & 96.00 & 35.67 & 30.33 & 91.67 & \textbf{85.79} & \textbf{43.00} & 65.08 & 64.17 & 40.33 & \underline{99.67} & 69.33 & 65.55 \\
+ FA-SSA Decode         & 97.00 & 37.67 & \underline{34.00} & \textbf{96.67} & 85.44 & 33.33 & \textbf{76.92} & 55.67 & \underline{50.33} & \textbf{100.00} & 72.00 & \textbf{67.19} \\
\midrule

\rowcolor{gray!10}
\multicolumn{13}{l}{\textit{Backbone: Qwen3-8B}} \\
Qwen3-8B                & 98.00 & 39.33 & \textbf{44.67} & 96.67 & 84.78 & \textbf{56.33} & 90.92 & 78.08 & \textbf{63.33} & \textbf{100.00} & 81.00 & \textbf{75.74} \\

\arrayrulecolor{black!40}\midrule
+ LycheeDecode          & 75.00 & 35.33 & 8.33  & 79.33 & 73.44 & 45.33 & 65.25 & 54.83 & 24.33 & \textbf{100.00} & 64.33 & 56.86 \\
+ TriangleMIX           & 98.00 & 40.13 & 16.33 & 97.33 & \underline{86.18} & 40.33 & \underline{91.17} & 79.10 & 49.33 & \textbf{100.00} & 77.33 & 70.47 \\
+ DuoAttention          & \underline{98.67} & 38.67 & 37.33 & 94.67 & 84.67 & 50.33 & 88.83 & 77.42 & 55.33 & \textbf{100.00} & 70.67 & 72.41 \\
+ PruLong               & 98.33 & 36.33 & 33.33 & 94.00 & 84.22 & 46.67 & 82.50 & 76.33 & 58.67 & \textbf{100.00} & 70.33 & 70.97 \\
+ XAttention            & 97.33 & 38.67 & 35.33 & 96.67 & 85.11 & 47.33 & 85.75 & 76.92 & 51.33 & \textbf{100.00} & \textbf{85.00} & 72.68 \\

\arrayrulecolor{black!40}\midrule
+ FA-SSA                & \underline{98.67} & \underline{42.00} & 20.00 & 97.00 & 85.89 & 50.67 & 90.58 & \underline{80.17} & \underline{59.00} & \textbf{100.00} & 79.33 & 73.03 \\
+ FA-XA                 & \textbf{99.00} & 39.00 & \underline{39.00} & 96.67 & 85.78 & \underline{52.00} & 87.92 & 78.75 & 58.67 & \textbf{100.00} & \underline{84.33} & \underline{74.65} \\
+ FA-TA                 & 97.33 & 39.33 & 33.33 & 97.33 & 85.56 & \underline{52.67} & \textbf{91.42} & 74.58 & 56.33 & \textbf{100.00} & 80.67 & 73.51 \\
+ FA-SSA Decode         & 97.33 & \underline{42.00} & 20.00 & \underline{97.67} & 84.33 & 51.00 & 88.92 & \textbf{82.00} & 56.00 & \textbf{100.00} & 78.33 & 72.51 \\
\midrule

\rowcolor{gray!10}
\multicolumn{13}{l}{\textit{Backbone: Llama-3.1-8B-Instruct}} \\
Llama-3.1-8B-Instruct            & \textbf{99.00} & \textbf{48.67} & \textbf{71.33} & 94.00 & 82.11 & \textbf{70.00} & \textbf{96.50} & \textbf{95.17} & 69.00 & \underline{99.67} & \textbf{92.67} & \textbf{83.47} \\

\arrayrulecolor{black!40}\midrule
+ LycheeDecode          & 34.67 & 43.67 & 20.00 & 77.33 & 69.78 & 56.33 & 67.50 & 66.08 & 38.67 & 98.67 & 77.00 & 59.06 \\
+ TriangleMIX           & 97.33 & \underline{48.33} & 62.00 & \underline{97.67} & 80.33 & 67.33 & 93.08 & 79.42 & \textbf{75.00} & 94.33 & 90.33 & 80.46 \\
+ DuoAttention          & 89.33 & 39.67 & 60.67 & 77.67 & 84.89 & 52.67 & 71.08 & 78.33 & 62.00 & 93.67 & 63.67 & 70.33 \\
+ PruLong               & 76.33 & 33.00 & 44.00 & 72.33 & 85.11 & 37.67 & 68.83 & 68.25 & 53.00 & 72.00 & 48.00 & 59.87 \\
+ XAttention            & 89.33 & 48.00 & 65.00 & 78.00 & 81.44 & 58.33 & 82.33 & 86.58 & 62.00 & \textbf{100.00} & 78.00 & 75.36 \\

\arrayrulecolor{black!40}\midrule
+ FA-SSA                & 98.33 & 42.67 & 59.33 & 95.00 & 84.44 & 41.33 & \underline{93.17} & 82.25 & 67.67 & 95.33 & 84.67 & 76.75 \\
+ FA-XA                 & 97.33 & \textbf{48.67} & \underline{68.33} & 88.00 & 82.33 & 60.33 & 88.75 & \underline{89.17} & 66.00 & \textbf{100.00} & 85.67 & 79.51 \\
+ FA-TA                 & \textbf{99.00} & 48.00 & 62.00 & \textbf{99.00} & 80.56 & \underline{69.00} & \underline{93.50} & 84.00 & \textbf{75.00} & 94.33 & \underline{92.00} & \underline{81.50} \\
+ FA-SSA Decode         & 90.00 & 44.40 & 70.80 & 86.40 & \textbf{86.27} & 47.20 & 90.30 & 71.00 & \underline{73.20} & 70.40 & 73.67 & 73.67 \\
\bottomrule
\end{tabular}
}
\end{table*}

Table~\ref{tab:ruler_refined} reports the tasks shared by the compared methods and omits Common Words Extraction (CWE) and Variable Tracking (VT). We complete the comparison for these two tasks in Table~\ref{tab:ruler_cwe_vt}: on Llama-3.1-8B-Instruct, Flux Attention improves the two-task aggregate over the dense backbone by 2.59 points and over DuoAttention by 0.35 points at approximately matched sparsity, and the conclusion holds across most of the evaluated length bins, with the advantage narrowing beyond 128K.
\begin{table}[t]
\centering
\caption{Common Words Extraction (CWE) and Variable Tracking (VT), the two RULER tasks omitted from Table~\ref{tab:ruler_refined}, evaluated on Llama-3.1-8B-Instruct under the same six context lengths $\{8\text{K}, 16\text{K}, 32\text{K}, 64\text{K}, 128\text{K}, 256\text{K}\}$. Including them preserves the main conclusion: FluxAttn remains ahead of both the dense backbone and the static head-level baseline at approximately matched sparsity. This added evaluation covers Llama-3.1-8B-Instruct only.}
\label{tab:ruler_cwe_vt}
\small
\setlength{\tabcolsep}{4pt}
\renewcommand{\arraystretch}{1.15}
\begin{tabular}{l c c c c}
\toprule
\textbf{Method} & \textbf{Sparsity} & \textbf{CWE} & \textbf{VT} & \textbf{Mean} \\
\midrule
Backbone (FA) & 0\% & 39.90 & 80.60 & 60.25 \\
FluxAttn & $\approx$51\% dynamic & 43.07 & 82.60 & \textbf{62.84} \\
DuoAttention & 50\% static & 42.10 & 82.87 & 62.49 \\
\bottomrule
\end{tabular}

\vspace{0.6em}

\begin{tabular}{l c c c}
\toprule
\textbf{Length} & \textbf{Backbone} & \textbf{FluxAttn} & \textbf{DuoAttention} \\
\midrule
0--8K & 58.0 & 64.8 & \textbf{66.0} \\
16K--32K & 61.6 & \textbf{81.2} & 78.8 \\
32K--64K & 85.2 & \textbf{86.0} & 77.2 \\
64K--128K & \textbf{96.0} & 94.0 & 93.2 \\
$>$128K & 91.4 & 84.8 & \textbf{91.0} \\
\bottomrule
\end{tabular}
\end{table}

\subsection{Sensitivity to the Sparse Attention Backend}
\label{appdix:backend_sensitivity}
Figure~\ref{fig:preliminary}(a) profiles sparsity sensitivity with Streaming Sparse Attention (SSA) under the configuration of \Cref{app:sparsity_config}. Because the observed degradation could in principle be an artifact of that backend, we repeat the sweep with XAttention (XA) and Triangle Attention (TA) on Llama-3.1-8B-Instruct under the same evaluation setting; Table~\ref{tab:backend_sensitivity} reports the per-category results normalized to the dense configuration.
Both backends reproduce the qualitative pattern of Figure~\ref{fig:preliminary}(a): retrieval-heavy document QA and multi-hop QA degrade sharply at high sparsity, while summarization and code tolerate aggressive sparsification. The onset and magnitude differ across backends, which is precisely why the Layer Router conditions its allocation on the chosen sparse branch instead of assuming a universal sparsity threshold. This is also the sense in which Flux Attention is backend-agnostic but not backend-indifferent: the framework routes at the layer level, and its measured benefits are reported for SSA, XA, and TA only. Methods such as Quest~\citep{quest}, MoBA, and DSA are natural candidates for the sparse branch, but we make no performance claim about them without an implementation and a matched experiment.
\begin{table*}[t]
\centering
\caption{Sensitivity of the Figure~\ref{fig:preliminary}(a) observation to the sparse operator, on Llama-3.1-8B-Instruct. Values are performance normalized to the dense configuration (100.00) and are averaged over the LongBench sub-tasks grouped below. Both XAttention (XA) and Triangle Attention (TA) reproduce the qualitative trend of Figure~\ref{fig:preliminary}(a)---retrieval-heavy QA degrades sharply at high sparsity while summarization and code remain tolerant---but the onset and magnitude of the degradation are backend-dependent, which is why the router conditions its allocation on the chosen sparse branch instead of assuming a universal threshold.}
\label{tab:backend_sensitivity}
\small
\setlength{\tabcolsep}{5pt}
\resizebox{\textwidth}{!}{
\begin{tabular}{c cccc cccc}
\toprule
\multirow{2}{*}{\textbf{Sparsity}} & \multicolumn{4}{c}{\textbf{\texttt{XAttention (XA)}}} & \multicolumn{4}{c}{\textbf{\texttt{Triangle Attention (TA)}}} \\
\cmidrule(lr){2-5} \cmidrule(lr){6-9}
 & Doc QA & Multi-Hop & Summ. & Code & Doc QA & Multi-Hop & Summ. & Code \\
\midrule
0.0 & 100.00 & 100.00 & 100.00 & 100.00 & 100.00 & 100.00 & 100.00 & 100.00 \\
0.2 & 102.78 & 96.26  & 99.97  & 98.54  & 101.76 & 99.10  & 94.86  & 95.63 \\
0.4 & 101.70 & 92.01  & 101.14 & 102.91 & 100.16 & 95.26  & 92.84  & 97.13 \\
0.6 & 96.87  & 92.37  & 101.04 & 104.54 & 97.98  & 93.21  & 92.32  & 97.01 \\
0.7 & 95.44  & 90.16  & 95.85  & 104.02 & 93.21  & 81.86  & 90.57  & 101.11 \\
0.8 & 89.71  & 88.33  & 96.02  & 102.00 & 87.20  & 83.65  & 88.05  & 96.14 \\
0.9 & 77.89  & 87.90  & 96.36  & 104.23 & 75.51  & 81.14  & 88.58  & 96.00 \\
\bottomrule
\end{tabular}
}
\end{table*}

\subsection{Router-Training Stochasticity}
\label{appdix:seed_variance}
Because the retained router is trained rather than calibrated, the headline LongBench-E comparison could in principle reflect a lucky initialization. Table~\ref{tab:seed_variance} reports two independently trained FA-SSA routers on Llama-3.1-8B-Instruct, the originally reported checkpoint and an additional run (seed 456). The aggregate difference is 0.08 points and no single task differs by more than 0.5 points, so the router-training noise is small relative to the reported gaps. Where the margin over the dense backbone is smaller than this band, we read the result as parity rather than as a reliable improvement, and we state the comparison accordingly.
\begin{table}[t]
\centering
\caption{LongBench-E per-task results for two independently trained FA-SSA routers on Llama-3.1-8B-Instruct. Router-training stochasticity shifts the task averages by at most 0.5 points, and the aggregate difference between the two runs is 0.08 points.}
\label{tab:seed_variance}
\small
\setlength{\tabcolsep}{10pt}
\begin{tabular}{l c c c}
\toprule
\textbf{Task} & \textbf{Original} & \textbf{Seed 456} & \textbf{Mean $\pm$ Std} \\
\midrule
Qasper          & 45.25 & 45.38 & 45.32 $\pm$ 0.07 \\
MF-en           & 54.42 & 54.42 & 54.42 $\pm$ 0.00 \\
HotQA           & 54.54 & 54.54 & 54.54 $\pm$ 0.00 \\
2Wiki           & 41.34 & 41.34 & 41.34 $\pm$ 0.00 \\
Gov.            & 34.54 & 34.71 & 34.63 $\pm$ 0.09 \\
M.News          & 26.16 & 26.22 & 26.19 $\pm$ 0.03 \\
TREC            & 68.33 & 68.74 & 68.54 $\pm$ 0.21 \\
TQA             & 91.91 & 91.92 & 91.92 $\pm$ 0.01 \\
SAMS            & 42.17 & 42.07 & 42.12 $\pm$ 0.05 \\
PCount          & 9.00  & 9.09  & 9.05 $\pm$ 0.05 \\
PRe             & 97.67 & 97.67 & 97.67 $\pm$ 0.00 \\
RB-P            & 47.74 & 48.24 & 47.99 $\pm$ 0.25 \\
Lcc             & 65.35 & 65.12 & 65.24 $\pm$ 0.12 \\
\arrayrulecolor{black!40}\midrule
\textbf{Avg.}   & 52.19 & 52.27 & 52.23 $\pm$ 0.04 \\
\bottomrule
\end{tabular}
\end{table}

\subsection{Batched Serving Throughput}
\label{appdix:batch_serving}
Single-request profiling can overstate the benefit of a sparsity scheme, because heterogeneous per-head workloads may be co-scheduled across requests once the server batches them. Table~\ref{tab:batch_serving} therefore measures aggregate decode throughput with concurrent requests at a 32K context. Flux Attention remains faster at every feasible batch size and reaches a larger feasible batch before running out of memory; we report the infeasible configurations as OOM rather than extrapolating throughput, and we state the claim for the tested serving setup rather than universally. At batch 32 both methods exhaust memory on this hardware.
\begin{table}[t]
\centering
\caption{Decode throughput under batched serving on Llama-3.1-8B-Instruct at a 32K context, measured on a node of 8$\times$NVIDIA H20 GPUs. Throughput is aggregate tokens/s across the served batch. OOM means the configuration exhausts GPU memory on this hardware; we report it rather than extrapolating throughput. The relative advantage does not disappear once heterogeneous requests are co-scheduled, and FluxAttn reaches a larger feasible batch before running out of memory.}
\label{tab:batch_serving}
\small
\setlength{\tabcolsep}{6pt}
\begin{tabular}{c c c c}
\toprule
\textbf{Batch size} & \textbf{FluxAttn (tok/s)} & \textbf{FA2 (tok/s)} & \textbf{FluxAttn / FA2} \\
\midrule
8  & 194 & 147 & 1.32$\times$ \\
16 & 196 & 95  & 2.07$\times$ \\
24 & 280 & OOM & --- \\
32 & OOM & OOM & --- \\
\bottomrule
\end{tabular}
\end{table}

\subsection{Full Quality--Latency--Memory Table}
\label{appdix:pareto_full}
Table~\ref{tab:pareto} gives the compact matched-sparsity comparison summarized in the main text, and Table~\ref{tab:pareto_full} the complete version with raw prefill and decode latencies, persistent KV memory, and peak memory per configuration, so that every reported speedup can be traced back to the measured quantities.
\begin{table*}[t]
\centering
\caption{Quality--latency--memory trade-off at matched sparsity on Llama-3.1-8B-Instruct. Accuracy is the RULER score per context length; speedups are relative to the dense FA2 baseline at the same context length, with prefill and decode reported separately because they are different phases. FluxAttn and PruLong operate at approximately matched model sparsity ($\Omega_{\mathrm{MSR}} \approx 0.50$), so the comparison isolates routing granularity rather than the amount of computation retained. ``KV saving'' is the reduction in persistent decode-time KV memory relative to dense. The accuracy entries coincide with the corresponding rows of Table~\ref{tab:ruler_lb2_result}, where the FluxAttn configuration is the sparse-decode one. Raw latencies and peak memory are given in Table~\ref{tab:pareto_full}, and the systems setup in Table~\ref{tab:systems_setup}.}
\label{tab:pareto}
\small
\setlength{\tabcolsep}{6pt}
\renewcommand{\arraystretch}{1.12}
\begin{tabular}{l c c c c c c}
\toprule
\textbf{Metric} & \textbf{8K} & \textbf{16K} & \textbf{32K} & \textbf{64K} & \textbf{128K} & \textbf{256K} \\
\midrule
Acc.\ Dense / full & 92.88 & 92.83 & 89.46 & 70.79 & 80.12 & 72.34 \\
Acc.\ PruLong / head & 86.96 & 76.55 & 70.65 & 54.52 & 48.18 & 30.00 \\
Acc.\ FluxAttn / layer & \textbf{90.11} & \textbf{79.39} & \textbf{79.22} & \textbf{56.08} & \textbf{62.94} & \textbf{59.39} \\
$\Omega_{\mathrm{MSR}}$ (PruLong / FluxAttn) & .50 / .53 & .50 / .50 & .50 / .53 & .50 / .53 & .50 / .50 & .50 / .50 \\
\arrayrulecolor{black!40}\midrule
Prefill speedup, PruLong & 1.06$\times$ & 1.14$\times$ & 1.25$\times$ & 1.42$\times$ & 1.60$\times$ & 1.76$\times$ \\
Prefill speedup, FluxAttn & 1.22$\times$ & 1.18$\times$ & 1.36$\times$ & 1.59$\times$ & 1.67$\times$ & 1.77$\times$ \\
\arrayrulecolor{black!40}\midrule
Decode speedup, PruLong & 1.15$\times$ & 1.23$\times$ & 1.27$\times$ & 1.36$\times$ & 1.43$\times$ & 1.51$\times$ \\
Decode speedup, FluxAttn & 1.19$\times$ & 1.30$\times$ & 1.41$\times$ & 1.53$\times$ & \textbf{1.77$\times$} & \textbf{2.22$\times$} \\
\arrayrulecolor{black!40}\midrule
KV saving, FluxAttn & 16.1\% & 28.9\% & 32.1\% & 37.8\% & 36.2\% & 39.3\% \\
\bottomrule
\end{tabular}
\end{table*}
\begin{table*}[t]
\centering
\caption{Full quality--latency--memory Pareto comparison at matched granularity controls on Llama-3.1-8B-Instruct; the compact view is Table~\ref{tab:pareto}. All methods are measured under the same precision and profiling procedure (Appendix~\ref{appendix:systems_setup}); speedups are relative to the dense FA2 baseline at the same context length. ``Persistent KV'' is the retained decode-time KV state and ``Peak'' the peak GPU memory. At 128K and 256K, where $\Omega_{\mathrm{MSR}}$ is exactly matched at $0.50$, FluxAtt improves accuracy over PruLong by 14.76 and 29.39 points, respectively, while reducing decode latency by 19.2\% and 32.0\% relative to PruLong. Prefill and decode columns report different inference phases and must not be conflated: head-level sparsity is competitive on prefill while remaining behind on per-token decode.}
\label{tab:pareto_full}
\small
\setlength{\tabcolsep}{4pt}
\renewcommand{\arraystretch}{1.1}
\resizebox{\textwidth}{!}{
\begin{tabular}{c l c c c c c c}
\toprule
\textbf{Ctx} & \textbf{Method / granularity} & \textbf{Acc.} & $\mathbf{\Omega_{\mathrm{MSR}}}$ & \textbf{Prefill ms (speedup)} & \textbf{Decode ms/tok (speedup)} & \textbf{Persistent KV GiB (saving)} & \textbf{Peak GiB} \\
\midrule
\multirow{3}{*}{8K}
 & Dense / full   & 92.88 & 0     & 985.0 (1.00$\times$)   & 25.17 (1.00$\times$) & 0.89 (0.0\%)  & 16.95 \\
 & PruLong / head & 86.96 & 0.500 & 926.2 (1.06$\times$)   & 21.88 (1.15$\times$) & 0.89 (0.0\%)  & 17.03 \\
 & FluxAtt / layer   & 90.11 & 0.531 & 807.4 (1.22$\times$)   & 21.15 (1.19$\times$) & 0.75 (16.1\%) & 16.89 \\
\arrayrulecolor{black!40}\midrule
\multirow{3}{*}{16K}
 & Dense / full   & 92.83 & 0     & 2280.1 (1.00$\times$)  & 16.31 (1.00$\times$) & 1.81 (0.0\%)  & 18.96 \\
 & PruLong / head & 76.55 & 0.500 & 2004.3 (1.14$\times$)  & 13.26 (1.23$\times$) & 1.81 (0.0\%)  & 19.09 \\
 & FluxAtt / layer   & 79.39 & 0.500 & 1932.3 (1.18$\times$)  & 12.55 (1.30$\times$) & 1.28 (28.9\%) & 18.29 \\
\arrayrulecolor{black!40}\midrule
\multirow{3}{*}{32K}
 & Dense / full   & 89.46 & 0     & 5918.8 (1.00$\times$)  & 19.97 (1.00$\times$) & 3.66 (0.0\%)  & 23.04 \\
 & PruLong / head & 70.65 & 0.500 & 4736.6 (1.25$\times$)  & 15.72 (1.27$\times$) & 3.66 (0.0\%)  & 23.27 \\
 & FluxAtt / layer   & 79.22 & 0.531 & 4352.1 (1.36$\times$)  & 14.16 (1.41$\times$) & 2.49 (32.1\%) & 21.46 \\
\arrayrulecolor{black!40}\midrule
\multirow{3}{*}{64K}
 & Dense / full   & 70.79 & 0     & 17201.4 (1.00$\times$) & 20.82 (1.00$\times$) & 7.39 (0.0\%)  & 31.07 \\
 & PruLong / head & 54.52 & 0.500 & 12146.8 (1.42$\times$) & 15.31 (1.36$\times$) & 7.39 (0.0\%)  & 31.51 \\
 & FluxAtt / layer   & 56.08 & 0.531 & 10818.5 (1.59$\times$) & 13.61 (1.53$\times$) & 4.60 (37.8\%) & 27.69 \\
\arrayrulecolor{black!40}\midrule
\multirow{3}{*}{128K}
 & Dense / full   & 80.12 & 0     & 54991.2 (1.00$\times$) & 32.42 (1.00$\times$) & 14.69 (0.0\%)  & 47.21 \\
 & PruLong / head & 48.18 & 0.500 & 34321.8 (1.60$\times$) & 22.67 (1.43$\times$) & 14.69 (0.0\%)  & 48.08 \\
 & FluxAtt / layer   & 62.94 & 0.500 & 32928.9 (1.67$\times$) & 18.32 (1.77$\times$) & 9.38 (36.2\%)  & 40.22 \\
\arrayrulecolor{black!40}\midrule
\multirow{3}{*}{256K}
 & Dense / full   & 72.34 & 0     & 190731.7 (1.00$\times$) & 54.57 (1.00$\times$) & 28.79 (0.0\%)  & 79.49 \\
 & PruLong / head & 30.00 & 0.500 & 108592.4 (1.76$\times$) & 36.14 (1.51$\times$) & 28.79 (0.0\%)  & 81.21 \\
 & FluxAtt / layer   & 59.39 & 0.500 & 107758.0 (1.77$\times$) & 24.58 (2.22$\times$) & 17.47 (39.3\%) & 65.28 \\
\arrayrulecolor{black}\bottomrule
\end{tabular}
}
\vspace{-0.5em}
\end{table*}

\subsection{Comparison with Elastic Attention}
\label{appendix:elastic_attn}

To ensure a fair evaluation, we establish a consistent baseline setting: both methods employ full attention for the retrieval mode and streaming attention for the sparse mode. For Elastic Attention~\citep{tang2026elasticattentiontesttimeadaptive}, we use its official implementation~\footnote{\url{https://github.com/LCM-Lab/Elastic-Attention}}. 

While Elastic Attention dynamically introduces sparsity at the head level to reduce theoretical FLOPs, this fine-grained approach often encounters hardware bottlenecks. Head-wise sparsity creates fragmented memory access patterns and computational load imbalance across GPU threads, limiting the actual wall-clock speedup. Flux Attention routes at the layer level, maintaining contiguous memory access for better hardware efficiency.

The two methods are also measured on different inference phases, and the numbers in Table~\ref{tab:elastic_comparison} are prefill/TTFT speedups. Elastic Attention's fused-kernel result likewise concerns prefill acceleration; our decode-phase measurements (Figure~\ref{fig:speedup}(b) and Table~\ref{tab:pareto}) show that the same granularity is markedly less effective per generated token, where historical-KV access and memory bandwidth dominate. The distinction matters when reading speedup claims: prefill speedup and per-token decode speedup are not interchangeable, and we annotate every system number in this paper by phase.

This efficiency is empirically validated in Table~\ref{tab:elastic_comparison}. Evaluated on Llama-3.1-8B-Instruct, Elastic Attention requires an aggressive sparsity of 68\% to achieve a 2.6$\times$ prefill speedup. Dropping this amount of context impairs its ability to process complex dependencies, reducing its RULER score to 72.85. Flux Attention operates at a lower sparsity of 51\%, i.e., it retains more attention computation. Due to its layer-wise efficiency, it still achieves a 2.5$\times$ prefill speedup. By preserving more context, Flux Attention improves the RULER score to 76.75 (+3.9) while maintaining comparable performance on LongBench.

\begin{table}[htbp]
\centering
\caption{Performance and efficiency comparison between Elastic Attention and Flux Attention on Llama-3.1-8B-Instruct. Flux Attention operates at a substantially lower sparsity level (51\% vs. 68\%), i.e., it retains more attention computation, and still achieves a comparable prefill/TTFT speedup (2.5$\times$ vs. 2.6$\times$) thanks to its hardware-friendly layer-wise routing. The reported speedups are prefill speedups; the decode-phase comparison is given in Table~\ref{tab:pareto}. Retaining more attention computation allows Flux Attention to deliver superior performance on the rigorous RULER benchmark while maintaining competitive results on LongBench.}
\label{tab:elastic_comparison}
\small
\setlength{\tabcolsep}{4.5pt} 
\renewcommand{\arraystretch}{1.15}
\begin{tabular}{l|c|c|cc|c}
\hline
\multirow{2}{*}{\textbf{Method}} & \multirow{2}{*}{\textbf{Granularity}} & \multirow{2}{*}{\textbf{Sparsity}} & \multicolumn{2}{c|}{\textbf{Performance}} & \multirow{2}{*}{\textbf{Speedup}} \\ \cline{4-5}
 & & & \textbf{LongBench} & \textbf{RULER} & \\ \hline
Elastic Attention & Head-wise & 68\% & \textbf{53.35} & 72.85 & \textbf{2.6$\times$} \\ 
\textbf{Flux Attention (Ours)} & \textbf{Layer-wise} & \textbf{51\%} & 52.28 & \textbf{76.75} & 2.5$\times$ \\ \hline
\end{tabular}
\end{table}

To verify the generalizability of our method, we extend the comparison across varying model architectures and scales, including Qwen3-4B, Qwen3-8B, and Llama-3.1-8B-Instruct. As shown in Table~\ref{tab:comprehensive_performance}, although Elastic Attention matches the dense performance on LongBench, it degrades significantly on RULER. For example, its Llama-3.1-8B-Instruct RULER score drops from 83.47 (dense baseline) to 72.85. 
Flux Attention consistently outperforms Elastic Attention in RULER across all evaluated models. In Qwen3-4B, our method surpasses Elastic Attention (66.95 vs. 61.81) and slightly exceeds the dense baseline. 

\begin{table}[htbp]
\centering
\caption{Comprehensive performance comparison among Dense Attention, Elastic Attention, and our Flux Attention across various foundation models. Flux Attention effectively mitigates the severe performance degradation seen in Elastic Attention on the RULER benchmark while maintaining highly competitive results on LongBench.}
\label{tab:comprehensive_performance}
\small
\setlength{\tabcolsep}{4.5pt} 
\renewcommand{\arraystretch}{1.15}
\begin{tabular}{l|l|cc}
\hline
\multirow{2}{*}{\textbf{Model}} & \multirow{2}{*}{\textbf{Method}} & \multicolumn{2}{c}{\textbf{Performance}} \\ \cline{3-4} 
 & & \textbf{LongBench} & \textbf{RULER} \\ \hline
\multirow{3}{*}{Qwen3-4B} 
 & Dense (Full Attention) & 48.45 & 66.00 \\ 
 & Elastic Attention & 48.08 & 61.81 \\ 
 & \textbf{Flux Attention (Ours)} & \textbf{48.72} & \textbf{66.95} \\ \hline
\multirow{3}{*}{Qwen3-8B} 
 & Dense (Full Attention) & 52.16 & 75.74 \\ 
 & Elastic Attention & 51.51 & 71.74 \\ 
 & \textbf{Flux Attention (Ours)} & \textbf{52.18} & \textbf{73.03} \\ \hline
\multirow{3}{*}{Llama-3.1-8B-Instruct} 
 & Dense (Full Attention) & 53.28 & 83.47 \\ 
 & Elastic Attention & \textbf{53.35} & 72.85 \\ 
 & \textbf{Flux Attention (Ours)} & 52.28 & \textbf{76.75} \\ \hline
\end{tabular}
\end{table}

\section{Analysis}
\label{appdix:abla_study}

\subsection{Impact of Data Composition on Task Differentiation}
\label{app:data_composition}

In previous sections, we have established that Flux Attention dynamically tailors sparsity to specific task demands. To fully unleash this capability, we discover that a well-balanced training curriculum acts as a crucial catalyst. To empirically validate this, we analyze the routing dynamics—specifically, the evolution of sparsity levels across training steps—under different data distribution settings.

\begin{figure*}[ht!]
    \centering
    \includegraphics[width=\linewidth]{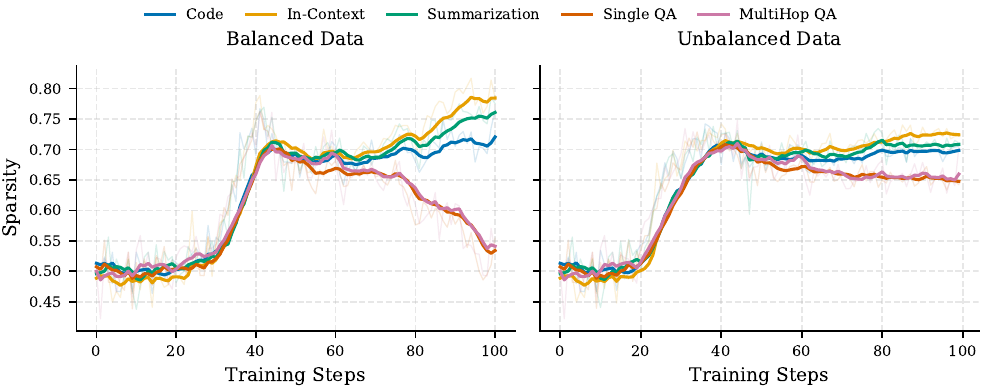}
    \caption{Evolution of sparsity levels across training steps under different data distributions. \textbf{Left:} Training on a well-balanced dataset, where the router successfully disentangles tasks into distinct sparsity levels. \textbf{Right:} Training on an unbalanced dataset dominated by context-holistic tasks, leading to homogenized routing.}
    \label{fig:spa_comparison}
\end{figure*}

Figure \ref{fig:spa_comparison} (Left) illustrates the sparsity trajectories when the router is trained on a well-balanced dataset. Driven by this diverse curriculum, the router successfully disentangles the underlying task demands and exhibits a clear divergence in its routing behavior. Notably, after an initial shared exploration phase, retrieval-intensive tasks converge to a lower sparsity level to preserve critical historical keys and values. In contrast, context-holistic tasks confidently sparsify the context, diverging toward higher sparsity levels. This demonstrates that a balanced mixture effectively teaches the router to establish robust, task-specific boundaries.

Conversely, Figure \ref{fig:spa_comparison} (Right) demonstrates the routing behavior when the training data is heavily skewed (e.g., dominated by context-holistic tasks). Under this setting, the router faithfully optimizes for the predominant data distribution. Rather than maintaining distinct task boundaries, the sparsity trajectories fail to clearly diverge after the initial phase, naturally converging toward a shared target sparsity. This results in a more homogenized routing strategy tailored to the specific domain it was exposed to.

This analysis yields an important insight into the training dynamics of the Layer Router: the router intrinsically aligns its allocation strategy with the global optimization landscape provided by the training data. Therefore, to train a general-purpose model capable of fine-grained, context-aware sparsification across diverse tasks, constructing a balanced task mixture during training is the optimal and highly effective practice.

\subsection{Impact of Input Truncation on Task Identification}
\label{appdix:truncation_analysis}

\begin{figure}[t]
\centering
\includegraphics[width=0.8\linewidth]{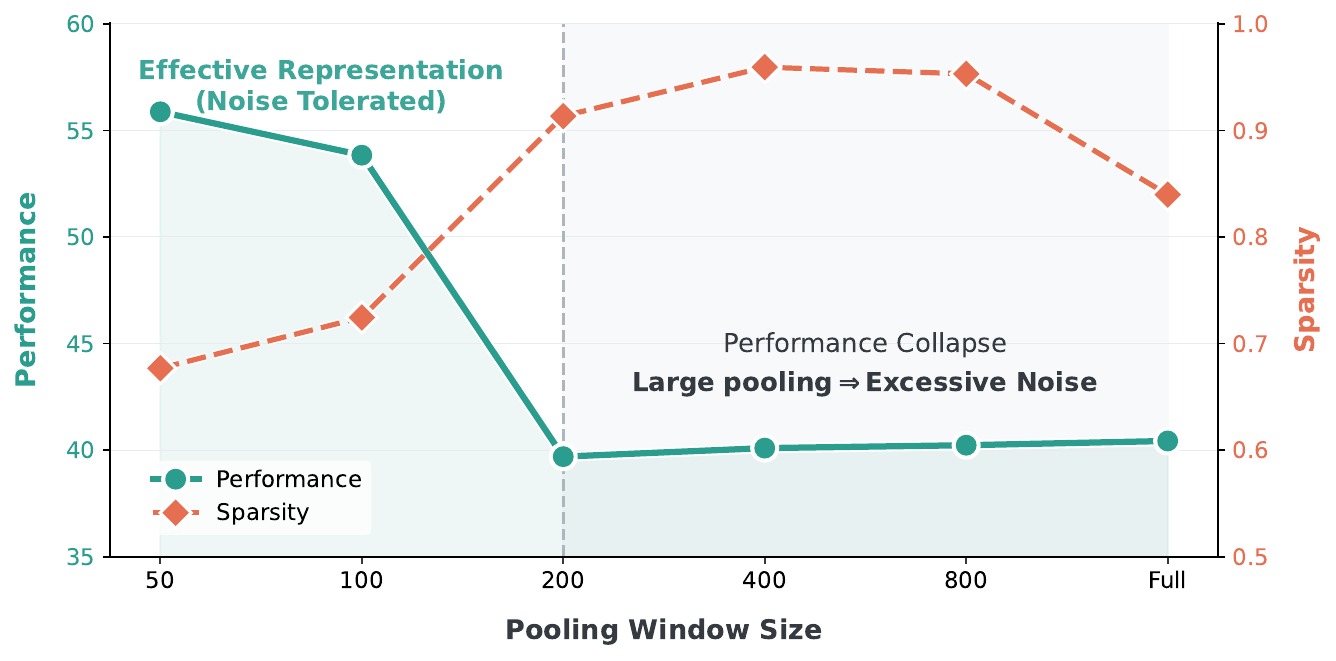}
\caption{
Impact of pooling window size on downstream performance and routing sparsity ($\Omega_{\mathrm{MSR}}$).
We evaluate varying truncation budgets ($L \in \{50, 100, 200, 400, 800, \text{Full}\}$), retaining only the sequence boundaries (prefix and suffix).
Increasing the pooling size beyond 100 tokens introduces context noise, which disrupts the routing mechanism. Consequently, the router misclassifies task features and assigns excessive sparsity to retrieval-intensive tasks, thereby degrading the overall performance.
}
\label{fig:truncation_analysis}
\end{figure}

To optimize the trade-off between routing efficiency and accuracy, we investigate the sensitivity of the layer router to the input sequence length. Specifically, we analyze how varying the truncation budget influences the capacity of the router to distinguish between task types and allocate appropriate sparsity patterns. Figure~\ref{fig:truncation_analysis} illustrates the performance and sparsity trends as the pooling window expands from 50 tokens (boundary-only) to the full sequence.

Our default strategy extracts only the first and last 100 tokens. This design leverages the structure of long-context prompts, where task-defining instructions typically appear at the beginning of the sequence, and specific user queries are appended at the end. The intermediate content primarily consists of raw context. Although this context is necessary for generation, it acts as noise during the routing process, which focuses on macro-level task identification.

Contrary to the assumption that additional context improves routing, Figure~\ref{fig:truncation_analysis} demonstrates a drop in performance when the pooling size exceeds 100 tokens.
We attribute this phenomenon to the limited capacity of the lightweight MLP within the routing module. As the pooling window expands, the task identification signals are diluted by the document tokens. The MLP struggles to filter out this noise and fails to capture the semantic features necessary for classification.
Consequently, the router makes suboptimal decisions, such as assigning high sparsity levels ($>0.9$) to retrieval-intensive tasks that require denser attention.
This misallocation causes the observed decrease in the quality of generation. These findings support the choice of a 100-token boundary window to maintain an optimal signal-to-noise ratio and facilitate accurate feature extraction.

\subsection{Relation Between Offline Entropy and Learned Routing}
\label{appdix:entropy_routing}
Section~\ref{subsec:observations} motivates layer-level routing with an offline entropy score $E_{\ell}$ (\Cref{appendix:layer_sparsification_layer}), which raises a natural question: does the trained router simply reproduce that static ranking? We measure this directly rather than assuming it.

$E_{\ell}$ is task-independent and computed once from a trace-normalized hidden-state Gram matrix using its top-128 eigenvalues on a profiling split of 26 disjoint, task-balanced LongBench-E samples. We then collect both soft and hard routing decisions over 20 samples from each of 13 LongBench-E tasks (260 samples in total, 32 layers) and correlate $E_{\ell}$ with each task's mean soft FA probability across layers. Table~\ref{tab:entropy_routing} reports the result.
The correlations are weak to moderate and task-dependent, ranging from $+0.027$ (Qasper) to $-0.314$ (MultiNews), with a macro-average Spearman correlation of $-0.215$ and a macro-average Pearson correlation of $-0.108$. The macro-average Spearman correlation is stable to the eigenvalue-truncation threshold ($-0.231$, $-0.234$, $-0.284$, $-0.215$ for $K = 16, 32, 64, 128$). The interpretation we adopt is deliberately limited: entropy is an offline indicator that layer sparsifiability is heterogeneous, not a routing label and not a supervision target. The router is trained from contextual representations and the task objective through the differentiable routing mechanism, so it remains free to depart from the static ordering, and it does so. We considered instead supervising the router directly with the static entropy ranking and chose not to: doing so would force a fixed layer ordering and could not express the input-dependent behavior observed above, where the same layer carries a near-zero FA probability on one task and a near-one probability on another.

The modest aggregate correlation is also not evidence that the router ignores the backbone's structure. Spearman correlation measures whether the layer \emph{ordering} is preserved, not the magnitude of the routing change: the mean cross-task correlation between routing patterns is $0.918$, so a shared architectural prior is retained, while a subset of layers moves substantially with the task. Layer 11, for instance, has a soft FA probability of $0.000$ on Qasper but $0.979$ on SAMSum and $0.984$ on RepoBench-P, and the overall FA share ranges from 35.2\% to 65.8\% across the 13 tasks. Large behavioral adaptation can therefore occur without completely reordering the layers.

\begin{table}[t]
\centering
\caption{Corrected correlation between the task-independent offline entropy score $E_{\ell}$ (Appendix~\ref{appendix:layer_sparsification_layer}) and the learned soft FA probability of each layer, on Llama-3.1-8B-Instruct. $E_{\ell}$ is computed once from a trace-normalized hidden-state Gram matrix using its top-128 eigenvalues on a profiling split of 26 disjoint task-balanced LongBench-E samples; routing is collected over 20 samples from each of 13 LongBench-E tasks (260 samples total) across 32 layers. Correlations are weak to moderate and task-dependent, so the learned router departs from a static entropy ranking rather than reproducing it; entropy is not used as a routing label or supervision target. The macro-average Pearson correlation is $r = -0.108$, and the macro-average Spearman correlation is stable to the eigenvalue-truncation threshold ($-0.231$, $-0.234$, $-0.284$, $-0.215$ for $K = 16, 32, 64, 128$).}
\label{tab:entropy_routing}
\small
\setlength{\tabcolsep}{10pt}
\begin{tabular}{l c c}
\toprule
\textbf{Task} & \textbf{Spearman $\rho$} & \textbf{Mean FA} \\
\midrule
Qasper            & $+0.027$ & 35.2\% \\
MultiFieldQA      & $-0.250$ & 44.2\% \\
HotpotQA          & $-0.205$ & 38.4\% \\
2WikiMQA          & $-0.167$ & 39.6\% \\
GovReport         & $-0.293$ & 57.5\% \\
MultiNews         & $-0.314$ & 57.7\% \\
TREC              & $-0.129$ & 41.7\% \\
TriviaQA          & $-0.175$ & 50.0\% \\
SAMSum            & $-0.237$ & 61.9\% \\
PassageCount      & $-0.278$ & 38.6\% \\
PassageRetrieval  & $-0.239$ & 46.8\% \\
LCC               & $-0.285$ & 65.4\% \\
RepoBench-P       & $-0.246$ & 65.8\% \\
\arrayrulecolor{black!40}\midrule
\textbf{Macro average} & $-0.215$ & 49.4\% \\
\bottomrule
\end{tabular}
\end{table}

\subsection{Routing Across Conversation Turns}
\label{appdix:multiturn}
Flux Attention routes once per prefill and reuses the cached decision for the whole decode trajectory, so a natural concern is whether the assignment becomes stale in agentic or multi-turn use, where the context can shift considerably during generation. Our design refreshes the route at the natural turn boundary: when the dialogue is extended, the enlarged history is re-prefilled and the router runs again, so the cached assignment is fixed only within one decode trajectory rather than globally. Within a response we deliberately do not reroute per token, because per-token rerouting would add control overhead and would break the homogeneous per-layer execution that produces the systems benefit, while the underlying task intent is usually stable over a single response.

To test that the refresh is substantive rather than nominal, we evaluated 80 two-turn MT-Bench conversations. After the first-turn interaction is appended to the context, 92.5\% of conversations changed at least one routing decision, and the average sparse-layer ratio moved from 45.8\% to 59.4\% (+13.6 points). The cached route is therefore trajectory-local: stable within a response, adaptive at turn boundaries.

\subsection{Composability with Fine-Grained Sparse Branches}
\label{appdix:composability}
Flux Attention selects the execution mode of a whole layer and does not constrain how the sparse branch is implemented. To check that this abstraction is usable rather than merely nominal, we integrated a vAttention-style~\citep{desai2026vattention} per-head sparse branch with Llama-3.1-8B-Instruct on an H20 GPU without modifying the router interface. The routed output stays close to full attention, with a cosine similarity of 0.984.

This result establishes functional composability at the attention interface, and we deliberately do not overstate it: it verifies integration and output agreement, not the wall-clock benefit of a fully optimized vAttention kernel under layer routing, because vAttention itself operates at a finer granularity than the layer. The relationship is complementary---Flux Attention chooses whether a layer runs FA or SA, while a fine-grained operator such as vAttention can then allocate sparsity inside the selected sparse branch. The same argument applies to the block-level methods we discuss in Section~\ref{sec:related_work}: Sketch-and-Walk~\citep{le2026sketchandwalk} and PBS-Attn~\citep{wang2026pbsattn} determine which token blocks are attended within a layer and could serve as the sparse branch in principle, which we describe as architectural compatibility rather than measured integration. Our measured wall-clock results cover SSA, XAttention, and Triangle Attention.

\subsection{Loss Curves and Performance Metrics}

\label{appdix:loss_curve_monitoring_metrics}

\begin{figure*}[ht]
\centering
\begin{subfigure}[t]{0.48\textwidth}
\centering
\includegraphics[width=\linewidth]{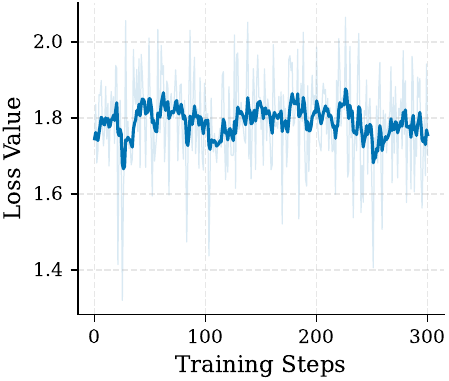}
\caption{Language Modeling Loss}
\label{fig:lm_loss}
\end{subfigure}
\hfill
\begin{subfigure}[t]{0.48\textwidth}
\centering
\includegraphics[width=\linewidth]{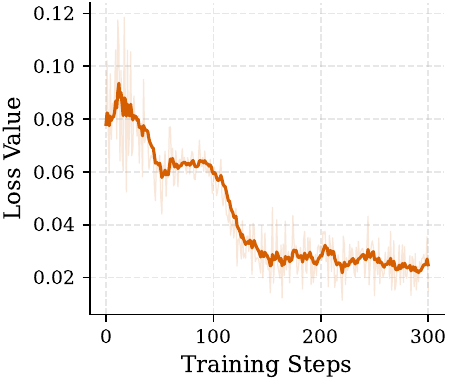}
\caption{Sparsity Regularization Loss}
\label{fig:reg_loss}
\end{subfigure}
\hfill
\begin{subfigure}[t]{0.48\textwidth}
\centering
\includegraphics[width=\linewidth]{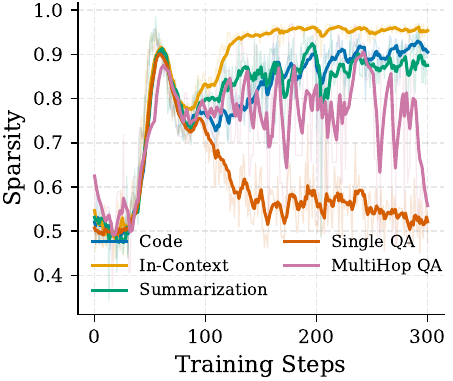}
\caption{$\Omega_{\mathrm{MSR}}$ during the training process}
\label{fig:sparsity_loss1}
\end{subfigure}
\hfill
\begin{subfigure}[t]{0.48\textwidth}
\centering
\includegraphics[width=\linewidth]{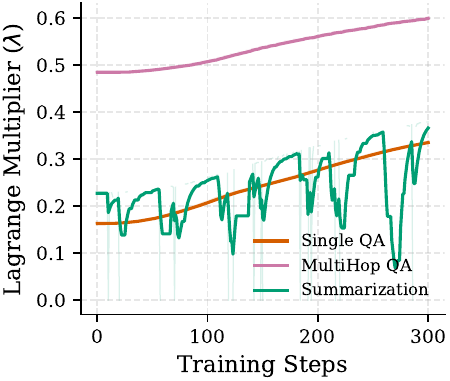}
\caption{Adaptive Coefficients ($\lambda$)}
\label{fig:sparsity_loss2}
\end{subfigure}
\caption{Decomposition of Training Objectives for Flux Attention. We visualize the training dynamics of the Layer Router, separating the total loss into (a) the primary language modeling objective and (b) the sparsity regularization term. Subfigures (c) and (d) illustrate the task-level differentiation in sparsity allocation ($\Omega_{\mathrm{MSR}}$) and adaptive coefficients ($\lambda$), demonstrating how the model automatically distinguishes between context-holistic  and retrieval-intensive tasks.}
\label{fig:lm_reg_loss}
\end{figure*}

We examine the training stability and dynamic routing behavior of Flux Attention by visualizing the optimization dynamics in Figure \ref{fig:lm_reg_loss}. This analysis decomposes the training process into the primary language modeling loss, the sparsity regularization loss, the evolution of the routed sparsity metric ($\Omega_{\mathrm{MSR}}$), and the adaptive coefficients ($\lambda$).

\paragraph{Optimization Stability.} As shown in Figures \ref{fig:lm_loss} and \ref{fig:reg_loss}, the joint optimization of the language modeling objective and Layer Router parameters remains stable. The LM loss decreases rapidly and plateaus around 1.8, suggesting that the lightweight Layer Router and the introduced sparsity do not impede convergence. Meanwhile, the sparsity regularization loss drops significantly within the first 100 steps. This indicates that the continuous relaxation scheme via Gumbel-Softmax effectively guides the router toward the specified sparsity constraints.

\begin{wrapfigure}{r}{0.5\linewidth}
  \centering
  \vspace{-1em}
  \includegraphics[width=\linewidth]{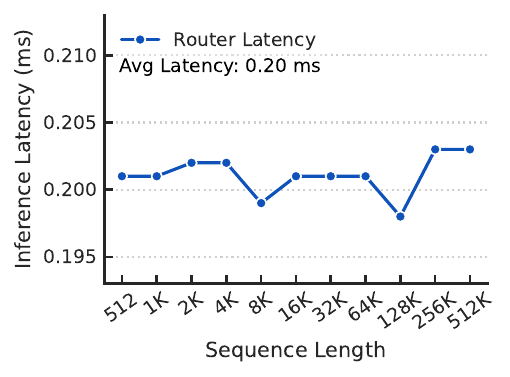}
  \caption{\textbf{Router latency analysis.} The router incurs negligible overhead (avg. 0.20 ms). Our design ensures length-invariant stability, maintaining constant speed from 512 to 1M tokens.}
  \label{fig:router_latency}
  \vspace{-1em}
\end{wrapfigure}

\paragraph{Differentiation in Flux Attention Allocation.} Figure \ref{fig:sparsity_loss1} provides empirical support for our motivation in Section \ref{sec:intro}, showing that downstream tasks exhibit varying sensitivities to attention sparsity. Starting from a neutral initialization, the Layer Router learns to differentiate between task types automatically. Retrieval-intensive tasks converge to lower $\Omega_{\mathrm{MSR}}$ values, i.e., they retain a larger allocation of Full Attention to preserve performance. In contrast, context-holistic tasks stabilize at higher values near the target threshold. Recall that under the convention of \Cref{equ:msr}, a higher $\Omega_{\mathrm{MSR}}$ means more Sparse Attention and less Full Attention. This confirms that Flux Attention identifies tasks capable of tolerating higher sparsity, thereby improving inference throughput without redundant computation.

\paragraph{Adaptive Coefficients.} Figure \ref{fig:sparsity_loss2} tracks the evolution of the Lagrangian multipliers ($\lambda$), which dynamically scale the penalty for sparsity violations. We observe that $\lambda$ increases most aggressively for certain tasks, suggesting the model prioritizes meeting density requirements where necessary. This adaptive mechanism balances the trade-off between computational cost and model quality automatically, eliminating the need for manual, task-specific tuning.

\section{Error Analysis}
\label{appdix:error_analysis}
In Table~\ref{fig:case_news}, \ref{fig:case_philosophical_argument}, and \ref{fig:case_paper_methodology}, we present representative model outputs comparing our method with other baselines. 
Due to the extensive length of the contexts, only a partial input context is shown. 
We observe that the primary source of performance improvement stems from our method's ability to accurately identify and respond to the key contextual segments relevant to the query.

\begin{figure}[ht]
\begin{AcademicBox}[Case 1: Reading Comprehension \& Information Extraction]
\small 
\setlength{\parindent}{0pt}

\textbf{\textit{Context:}} \\
"Low-cost solutions can give billions access to modern cooking by 2030, but the world is failing to deliver...
Nearly one in three people around the world still cook their meals over open fires or on basic stoves, resulting in significant damage to health, living standards and gender equality – and yet this challenge can be overcome this decade through a relatively modest amount of investment...
Today, 2.3 billion people rely on charcoal, firewood, coal, agricultural waste and animal dung as fuel to prepare meals, causing them to breathe in harmful smoke in the process..."
\vspace{4pt} \hrule \vspace{4pt}

\textbf{\textit{Question:}} \\
Based on the information provided in these news articles, which of the following conclusions regarding African clean cooking is the most reliable?
\vspace{4pt} \hrule \vspace{4pt}

\textbf{\textcolor{teal}{Correct Prediction (Ours / Ground Truth):}} \hfill \textcolor{teal}{\checkmark} \\
\textbf{Option B} \\
Content: \textbf{In comparison to other regions of the world, the issue of outdated cooking methods in Africa is particularly severe}: nearly four-fifths of the African population still rely on traditional stoves or open fires for cooking, a figure that stands at less than one half globally.
\vspace{4pt} \hrule \vspace{4pt}

\textbf{\textcolor{red}{Incorrect Baselines:}} \\
\footnotesize
\begin{itemize}[leftmargin=*, label={}, itemsep=4pt, topsep=2pt]

    \item \textbf{Base Model (Qwen3-4B), PruLong, DuoAttention, Triangle}: \\
    Pred: Option A \\
    Content: \textbf{African nations broadly benefit from carbon markets, particularly in addressing financing issues for clean cooking}: Carbon credits can bridge the funding gap for clean cooking investments in Africa, while also enhancing the affordability of clean cookstoves and fuels. \\
    \textit{(Error: Baselines hallucinate or improperly infer information about carbon markets, failing to ground their conclusion strictly in the provided text's severity statistics.)}

\end{itemize}

\end{AcademicBox}
\caption{\textbf{Comparison on a long-context reading comprehension task.} Our model accurately extracts and verifies the severity statistics of outdated cooking methods in Africa compared to global figures, while all baselines consistently fall for the same unsupported distractor regarding carbon markets.}
\label{fig:case_news}
\end{figure}
\begin{figure}[ht]
\begin{AcademicBox}[Case 2: Core Argument Identification in Abstract Text]
\small 
\setlength{\parindent}{0pt}

\textbf{\textit{Context:}} \\
"The Lawyer as Friend: The Moral Foundations of the Lawyer-Client Relation'
Charles Friedt
Advocatus sed non ladro,
Res miranda populo ....
Medieval anthem honoring St. Ives

Can a good lawyer be a good person? The question troubles lawyers and law students alike. They are troubled by the demands of loyalty to one's client and by the fact that one can win approval as a good, maybe even great, lawyer even though that loyalty is engrossed by over-privileged or positively distasteful clients. How, they ask, is such loyalty compatible with that devotion to the common good characteristic of high moral principles? And whatever their views of the common good, they are troubled because the willingness of lawyers to help their clients use the law to the prejudice of the weak or the innocent seems morally corrupt..."
\vspace{4pt} \hrule \vspace{4pt}

\textbf{\textit{Question:}} \\
What is the core argument of this article? \\
\vspace{4pt} \hrule \vspace{4pt}

\textbf{\textcolor{teal}{Correct Prediction (Ours / Ground Truth):}} \hfill \textcolor{teal}{\checkmark} \\
\textbf{Option C} \\
Content: \textbf{Refuting the social doubts about lawyers' professional ethics through analogy.}
\vspace{4pt} \hrule \vspace{4pt}

\textbf{\textcolor{red}{Incorrect Baselines:}} \\
\footnotesize
\begin{itemize}[leftmargin=*, label={}, itemsep=4pt, topsep=2pt]
    \item \textbf{Base Model (Llama3.1-8B), DuoAttention, Triangle}: \\
    Pred: Option B \hspace{10pt} Content: \textbf{A good lawyer can be a good person.} \\
    \textit{(Error: Baselines fail to abstract the overarching argumentative framework, instead parroting the literal opening rhetorical question as the core argument itself.)}

    \item \textbf{PruLong}: \\
    Pred: Option A \hspace{10pt} Content: \textbf{Lawyers should be regarded as friends of clients.} \\
    \textit{(Error: The model superficially extracts the title text ("The Lawyer as Friend") without understanding that the "friend" concept is merely the analogy used to resolve the broader ethical doubts.)}

\end{itemize}

\end{AcademicBox}
\caption{\textbf{Qualitative comparison on identifying the core argument in a philosophical legal text.} Our model successfully synthesizes the text to identify the underlying argumentative strategy (refutation via analogy), whereas baselines are easily distracted by literal sentences from the title and opening hook.}
\label{fig:case_philosophical_argument}
\end{figure}
\begin{figure}[ht]
\begin{AcademicBox}[Case 3: Methodology Extraction from Academic Paper]
\small 
\setlength{\parindent}{0pt}

\textbf{\textit{Context:}} \\
"Published as a conference paper at ICLR 2024
MAGICDRIVE: STREET VIEW GENERATION WITH DIVERSE 3D GEOMETRY CONTROL
ABSTRACT
Recent advancements in diffusion models have significantly enhanced the data synthesis with 2D control. Yet, precise 3D control in street view generation, crucial for 3D perception tasks, remains elusive. Specifically, utilizing Bird’s-Eye View (BEV) as the primary condition often leads to challenges in geometry control (e.g., height), affecting the representation of object shapes, occlusion patterns, and road surface elevations... In this paper, we introduce MAGICDRIVE, a novel street view generation framework, offering diverse 3D geometry controls including camera poses, road maps, and 3D bounding boxes, together with textual descriptions, achieved through tailored encoding strategies..."
\vspace{4pt} \hrule \vspace{4pt}

\textbf{\textit{Question:}} \\
How does the MagicDrive encode the bounding box in training step? \\
\vspace{4pt} \hrule \vspace{4pt}

\textbf{\textcolor{teal}{Correct Prediction (Ours / Ground Truth):}} \hfill \textcolor{teal}{\checkmark} \\
\textbf{Option D} \\
Content: \textbf{MagicDrive encode the class labels and the corner points separately, and then uses an MLP to compress them into a vector.}
\vspace{4pt} \hrule \vspace{4pt}

\textbf{\textcolor{red}{Incorrect Baselines:}} \\
\footnotesize
\begin{itemize}[leftmargin=*, label={}, itemsep=4pt, topsep=2pt]
    \item \textbf{Base Model (Qwen3-8B), PruLong, DuoAttention, Triangle}: \\
    Pred: Option B \hspace{10pt} Content: \textbf{MagicDrive uses Fourier embedding for 4 corner point and passes them through an MLP for encoding in training step. MagicDrive then uses an MLP to compress both the class label and position embeddings into a single hidden vector for each bounding box.} \\
    \textit{(Error: Baselines hallucinate or misattribute specific technical details, such as "Fourier embedding for 4 corner points," failing to accurately extract the exact encoding mechanism described in the paper.)}

\end{itemize}

\end{AcademicBox}
\caption{\textbf{Qualitative comparison on extracting technical methodology from a machine learning paper.} Our model accurately identifies the specific bounding box encoding strategy, whereas all baselines suffer from hallucination, confidently generating plausible but incorrect architectural details (Fourier embeddings) not supported by the text.}
\label{fig:case_paper_methodology}
\end{figure}


\end{document}